\documentclass{article}

\usepackage{graphicx} 
\usepackage{authblk}
\usepackage{outlines}
\usepackage{parskip}
\usepackage{amsmath}
\usepackage{amsthm}
\usepackage{amssymb}
\usepackage{amsfonts}
\usepackage{hyperref}
\usepackage[normalem]{ulem} 
\usepackage{svg} 
\usepackage{subcaption} 
 \usepackage{dsfont} 
\usepackage{xcolor}
\usepackage{tikz}
\usepackage{tikz-3dplot}
\usetikzlibrary{arrows, calc, backgrounds, shapes.geometric, decorations.pathreplacing, angles, quotes}
\tdplotsetmaincoords{70}{20}
\usetikzlibrary{arrows.meta,positioning,calc}
\usepackage{float}
\hypersetup{colorlinks=true, unicode=true, linkcolor=[rgb]{0.10,0.05,0.67}, citecolor=[rgb]{0.10,0.05,0.67}, filecolor=[rgb]{0.10,0.05,0.67}, urlcolor=[rgb]{0.10,0.05,0.67}}
\usepackage[a4paper, total={6in, 8in}]{geometry} 
\newcommand{\inc}{\mathrel{\rotatebox[origin=c]{270}{$\curlyvee$}}}
\def\ORIENT{30}
\definecolor{skyblue}{RGB}{135,206,235}
\DeclareMathOperator*{\argmin}{arg \text{  } min} 

\usepackage{import} 
\usepackage[
  sorting=nyt,
  maxbibnames=99,
  minbibnames=99
]{biblatex}

\title{Book Chapter - AI + Defense}

\author[1]{Sarah Harkins Dayton\textsuperscript{*}}
\author[1]{Layal Bou Hamdan\thanks{These authors contributed equally to this work.}}

\author[2]{Ioannis D. Schizas}
\author[2]{David L. Boothe}
\author[1]{Vasileios Maroulas}
\affil[1]{Department of Mathematics, University of Tennessee, Knoxville}
\affil[2]{U.S. Army DEVCOM Army Research Laboratory}

\title{From Classical to Topological Neural Networks Under Uncertainty}

\date{}

\begin{document}

\maketitle

\begin{abstract}
    This chapter explores neural networks, topological data analysis, and topological deep learning techniques, alongside statistical Bayesian methods, for processing images, time series, and graphs to maximize the potential of artificial intelligence in the military domain. Throughout the chapter, we highlight practical applications spanning image, video, audio, and time-series recognition, fraud detection, and link prediction for graphical data, illustrating how topology-aware and uncertainty-aware models can enhance robustness, interpretability, and generalization.        
\end{abstract}







    \section{Introduction}

To remain the most powerful global force, the military must convert raw data into confident decisions faster than any adversary  \cite{GAO-25-106454,DoD2022JADC2Summary}. Artificial intelligence (AI) is a field focused on training machines to replicate human brain functions and perform intelligent tasks efficiently \cite{russel2009ai}. Modern AI compresses sensing, understanding, and action into seconds, making its implementation not just an advantage but a necessity. Operating at the speed of data, AI drives the mission, from long-range strategic planning to day-to-day command and control operations, while also increasing efficiency in logistics and maintenance. Specifically, AI implementation can assist with predictive maintenance to keep vehicles mission-ready \cite{Stanton2023PdMAircraft}, multi-sensor fusion to improve target recognition and battle damage assessment \cite{s23042207}, adaptive spectrum management to perform effective counter jamming and achieve electromagnetic spectrum dominance \cite{Zhang2025intelligent}, and threat-aware routing that adapts as the conditions evolve \cite{drones9050376}.
 
AI is built on artificial neural networks (ANN) that are biologically inspired computational models with wide applications. The design of ANNs is a formidable research frontier in many disciplines \cite{castillo2025artificial, mackey2025geometric, walczak2025eden, deighton2025functional} to address problems ranging from optimization, prediction, pattern recognition, and object classification \cite{ANN_intro}. 

In an ANN, there are three types of layers: (1) input layer, (2) hidden layers, and (3) output layer. The input layer receives information from the input data and passes it to the hidden layers. Each neuron in the input layer corresponds to a feature in the input data. The hidden layers perform the majority of the work of a neural network. Each hidden layer applies an assigned transformation to the input data, using a set of weights and biases, and then applies a nonlinear activation function. Nonlinear activation functions further increase the expressiveness of the network \cite{Kriegeskorte_2019}. Additional hidden layers may be incorporated to extract higher-order relationships out of the data, where fewer layers may be insufficient. The output layer produces the final output of the network. The number of neurons in an output layer corresponds to the number of classes in a classification problem, or the desired dimension of the output for a regression problem. 

\begin{figure}[h]
    \centering
    \includegraphics[height=1.2in]{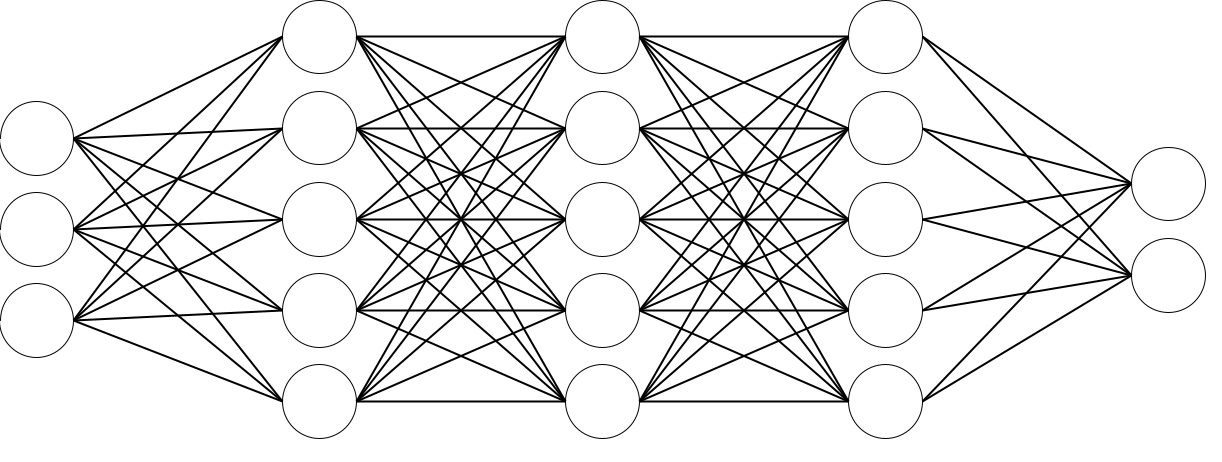}
    \caption{General architecture of an artificial neural network.}
    \label{fig:ann}
\end{figure}

There are two major groups of neural network architecture based on the connection styles: (1) feed-forward networks, in which information passes only in one direction, and (2) recurrent networks, in which information may be passed in a cyclic, or looping fashion \cite{AnOverviewofNeuralNetwork}. These two groups generally exhibit different network behavior. For instance, feedforward networks are considered memory-less since their outputs are independent of the previous network state, whereas recurrent networks consider previous network states with their cyclic connections, allowing them to remember information from previous time steps. 


In the realm of military operations, data is often drawn from heterogeneous, noisy, incomplete, and high-dimensional sources. Extracting reliable insights from such complex information streams poses a significant analytical challenge \cite{mitra2024usegraphneuralnetworks}. Topological data analysis (TDA) provides a powerful mathematical framework for uncovering the underlying structure and geometry of data, allowing analysts to detect patterns that are hidden by conventional methods \cite{Chazal2017AnIT,Carlsson2021tda,ghrist2008barcodes}. Although TDA focuses on the analysis of the shape of data to extract structural insights, recent work extends these ideas to topology-driven deep learning (TDL), which integrates topological principles directly into neural network architectures \cite{hofer2017deep}. This shift goes beyond the use of topology only as a post-analysis tool and instead embeds topological awareness within the learning process, allowing networks to preserve global structure, improve generalization, and adapt topological features during training \cite{hofer2017deep,bodnar2021simplicial,love2023tcnn,hansen2020sheaf,hajij2020cell,MitchellSpikeDecoding}.

Bayesian neural networks (BNNs) extend deep learning frameworks by introducing a probabilistic treatment of model parameters to enhance model interpretability and robustness \cite{blundell2015weightuncertaintyneuralnetworks,gal2016dropoutbayesianapproximationrepresenting,NEURIPS2020_322f6246,neal2012bayesian}. Instead of assigning fixed values to weights and biases, BNNs represent them as random variables with associated probability distributions. This probabilistic view captures epistemic uncertainty that arises from limited data or model capacity to provide a principled way to quantify confidence in learned features and predictions.

Recently, Bayesian formulations have integrated uncertainty estimation into topological representations to enable more reliable characterization of structural patterns in data, particularly in settings where noise, sparsity, or limited samples affect topological summaries \cite{maroulas2020abayesian,maroulas2022bayesian,gillespie2025bayesian,oballe2022discrete,NEURIPS2020_322f6246}. This combination supports tasks such as topological feature selection, manifold learning, and persistent homology inference under uncertainty, reinforcing the theoretical link between probabilistic modeling and topology-based learning frameworks.

This chapter presents a range of techniques from neural networks to Bayesian methods to topological data analysis and topological deep learning, and their use cases for handling images, time series, and graphs to maximize AI's potential in the military domain. It emphasizes how matching cutting-edge methods with specific forms of data produces measurable and repeatable advantages for applications in different disciplines, including military operations. It serves to underscore the importance of careful data curation, model calibration, and uncertainty management. By advancing human–machine teaming and accelerating the transition from experimental prototypes to operational capabilities, organizations can maintain superiority in an era where technological evolution defines success. The chapter is organized in two sections where the first section introduces core artificial neural networks and Bayesian methods, while the second addresses topological methods in neural networks.

    \section{Foundational Artificial Neural Networks}

\subsection{Convolutional Neural Networks}\label{sec:CNN}

\begin{figure}[h]
    \centering
    \includegraphics[width=0.85\linewidth]{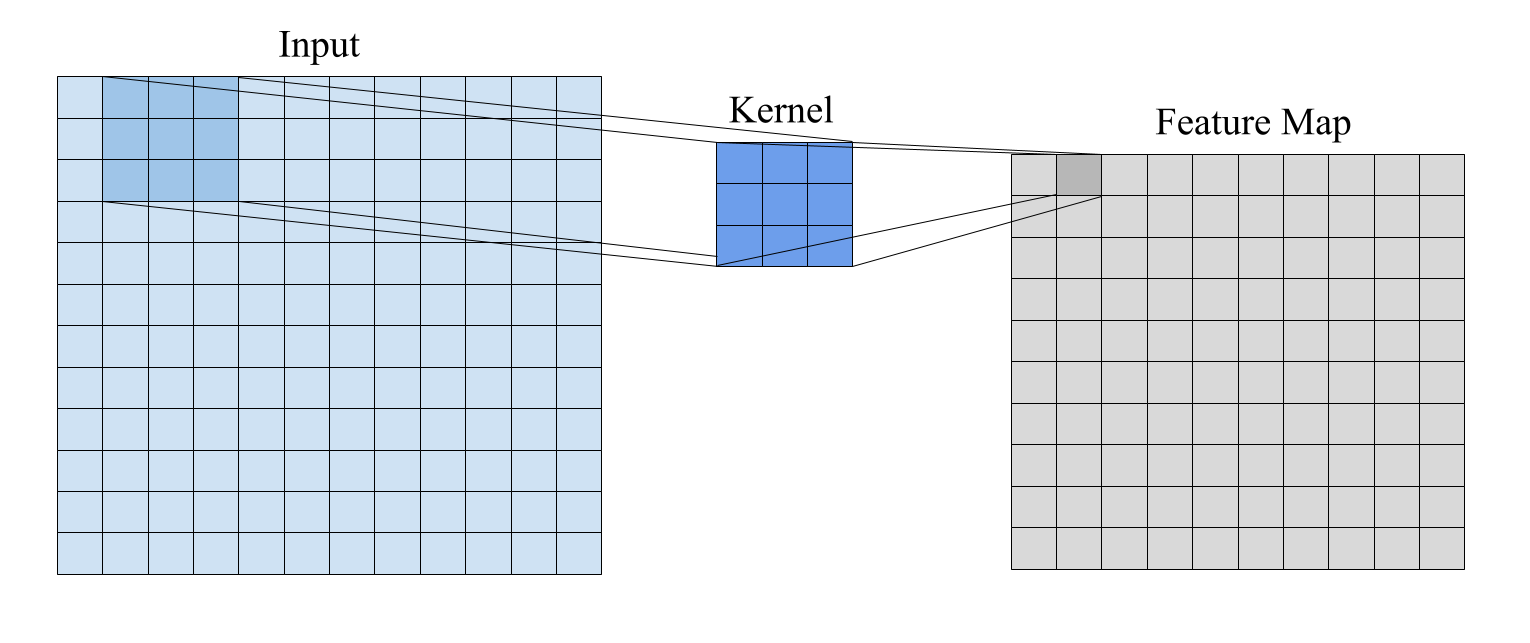} 
    \caption{A convolution operation that is applied in a convolutional layer.}
    \label{fig:convolution}
\end{figure}

Convolutional neural networks (CNNs) are a widely used machine learning technique for problems involving computer vision introduced by Yann LeCun, dubbed LeNet-5 \cite{LeNet}.  CNNs are a subset of ANNs that take inspiration from the neurons of a cat's visual system \cite{Hubel_Wiesel}. CNNs are generally comprised of three layer types: convolutional layers, pooling layers, and fully connected layers \cite{basic_cnn}. In the convolutional layer, a discrete convolution is applied where a sliding window, called a kernel or filter, moves across the data and computes a dot product between values in the kernel and values in the data at each position to create a feature map, as seen in Figure \ref{fig:convolution}. Formally, a discrete convolution on a one-channel image with a single kernel is defined by 
\begin{equation}
    (\mathbf{X} \star \mathbf{K})[i,j] = \sum_{m} \sum_{n}  \mathbf{X}[i-m, j-n] \mathbf{K}[m,n],
\end{equation}
where $(\mathbf{X} \star \mathbf{K})[i,j]$ is the $(i,j)$ position of the resulting feature map, $\mathbf{X}$ is the input image, and $\mathbf{K}$ is the kernel.

Multiple kernels can be used to create multiple feature maps within each convolutional layer. The goal of a convolution is to extract pertinent feature information from the data. The values in the kernel are learned during the training process. Pooling layers typically follow after convolutional layers to further extract feature information, mitigate redundancy in obtained features, and reduce the dimension of the feature maps. Max pooling \cite{avg_pooling} and average pooling \cite{LeNet} are typical pooling operations. In both of these pooling techniques, a pooling filter acts as a sliding window, similar to that in the convolution operation, on the feature map produced by the convolution operation. Pooling, in general, applies an action to the entries of the convolutional feature map, and the result is the entry of a new feature map. Max pooling takes the maximum of the entries in the pooling area. Average pooling takes the average of the entries in the pooling area. Pooling reduces the dimensionality of the feature maps in a network \cite{gholamalinezhad2020poolingmethodsdeepneural}. For image classification problems, once the features are extracted with convolutional layers, fully connected layers are typically used to perform the classification by mapping to the output classes.

\textbf{Applications:} CNNs are a highly capable foundational model for image and video processing. CNN-based models are capable of supporting many systems for target detection, entity recognition, and reconnaissance technology. In \cite{CNN_IR}, the authors utilize a CNN-based model for target recognition and classification via infrared images of vehicles like main battle tanks, armored personnel carriers, and pickup trucks. This work shows the robustness of a CNN-based model when the input images are perturbed due to inconsistencies when an entity is detected, leading to translated and scaled input images. In \cite{s23042207}, the authors used convolutional-based neural network techniques to fuse airborne radar and optical imagery to identify military vehicles and improve situational awareness. They implemented convolutional neural networks, including a ResNet-18 variant \cite{He2016resnet}, and showed that combining the two sensor types outperformed using either sensor alone, with a simple input-channel fusion scheme delivering the best results.

\subsection{Recurrent Neural Networks}
Recurrent neural networks (RNNs) are a category of neural networks that utilize cyclic connections between layers. The cyclic connections in the network allow for previous time steps to influence the current output. This consideration makes RNNs better suited to handle sequential data compared to feedforward neural networks \cite{Rumelhart_etal}. To create cyclic connections, RNNs compute a hidden state that considers a nonlinear activation applied to a linear combination of the previous hidden state and the current input. The hidden state for time step $j$ of an RNN is defined as 
\begin{equation}
    \mathbf{h}_{j} = \sigma(\mathbf{W}_{ih} \cdot \mathbf{x}_{j} + \mathbf{W}_{hh} \cdot \mathbf{h}_{j-1}+ \mathbf{b}_{h}),
\end{equation}
where $\mathbf{W}_{ih}, \mathbf{W}_{hh}$ are the respective weight matrices $\mathbf{x}_{j}$ is the input from time step $j$, $\mathbf{h}_{j-1}$ is the previous hidden state at time step $j-1$, $\mathbf{b}_{h}$ is the bias, and $\sigma(\cdot)$ is a nonlinear activation function. The weight matrix $\mathbf{W}_{ih}$ controls how much the current input $\mathbf{x}_{j}$ impacts the current hidden state $\mathbf{h}_{j}$, and $\mathbf{W}_{hh}$ determines how much the previous hidden state $\mathbf{h}_{j-1}$ impacts the current hidden state $\mathbf{h}_{j}$. RNNs are powerful architectures for capturing relationships in sequential data, but they tend to struggle with longer sequences due to vanishing and exploding gradients in their training processes \cite{Gers_Schmidhuber_Cummins_2000}.

Long short-term memory units (LSTMs) \cite{LSTM} were created as an extension of RNNs to combat vanishing and exploding gradients in RNNs. Along with a hidden state and cell state, LSTMs utilize a gating scheme that allows for learning of longer-term dependencies. The gating scheme is as follows for time step $j$:
\begin{equation}
\begin{aligned}
    \text{Input Gate: } & \mathbf{i}_{j} = \sigma_{i}\big(\mathbf{W}_{i} \mathbf{x}_{j} + \mathbf{U}_{i} \mathbf{h}_{j-1} + \mathbf{b}_{i}\big)\\
    \text{Forget Gate: } & \mathbf{f}_{j} = \sigma_{f}\big(\mathbf{W}_{f} \mathbf{x}_{j} + \mathbf{U}_{f} \mathbf{h}_{j-1} + \mathbf{b}_{f}\big)\\
    \text{Output Gate: } & \mathbf{o}_{j} = \sigma_{o}\big(\mathbf{W}_{o} \mathbf{x}_{j} + \mathbf{U}_{o} \mathbf{h}_{j-1} + \mathbf{b}_{o}\big)
\end{aligned}
\end{equation}
where $\mathbf{W}_{*}, \mathbf{U}_{*}$ are the respective weight matrices, $\mathbf{x}_{j}$ is the input at time step $j$, $\mathbf{h}_{j-1}$ is the previous hidden state are time step $j-1$, and  $\mathbf{b}_{*}$ is the bias, where the subscript $*$ refers to the input gate $i$, the forget gate $f$, or the output gate $o$. The input gate controls the usage of new information from the current input in the intermediate cell state. The forget gate determines how much of the precious state is preserved. The output gate generates a scaling factor that influences the calculation of the hidden state. 

In addition to the hidden state, LSTMs use an internal cell state, $\mathbf{c}_{j}$. First an intermediate cell state, $\tilde{\mathbf{c}}_{j}$, is computed as follows $\tilde{\mathbf{c}}_{j} =  \sigma_{c} (\mathbf{W}_{c} \cdot \mathbf{x}_{j} + \mathbf{U}_{c} \cdot \mathbf{h}_{j-1} + \mathbf{b}_{c}),$ where $\sigma_{c}$ is a nonlinear activation function, $\mathbf{W}_{c}, \mathbf{U}_{c}$ are the respective weight matrices, and $\mathbf{b}_{c}$ is the bias. The cell state, $\mathbf{c}_{j}$, is computed $\mathbf{c}_{j} = \mathbf{i}_{j} \odot \tilde{\mathbf{c}}_{j} +  \mathbf{c}_{j-1} \odot \mathbf{f}_{j},$ where $\mathbf{c}_{j-1}$ is the cell state at the previous time step $j-1$, where $\odot$ represents entrywise multiplication. Lastly, the hidden state is computed $\mathbf{h}_{j} = \mathbf{o}_{j}  \odot  \text{tanh} (\mathbf{c}_{j}).$

Although LSTMs address the vanishing and exploding gradients in RNNs, the intricate architecture of LSTMs makes it costly to train \cite{LSTM_downfalls}. LSTMs are also susceptible to overfitting on small datasets.

Gated recurrent units (GRUs), first proposed by \cite{cho2014learningphraserepresentationsusing}, simplify the gating scheme of LSTMs to alleviate the computational costs associated with training an LSTM. The GRU gating scheme is as follows for time step $j$:
\begin{equation}
    \begin{aligned}
    \text{Reset Gate: } & 
     \mathbf{r}_{j} = \sigma(\mathbf{W}_{r} \cdot \mathbf{x}_{j} + \mathbf{U}_{r} \cdot \mathbf{h}_{j-1}+ \mathbf{b}_{r}) \\
    \text{Update Gate: } & \mathbf{u}_{j} = \sigma(\mathbf{W}_{u} \cdot \mathbf{x}_{j} + \mathbf{U}_{u} \cdot \mathbf{h}_{j-1}+ \mathbf{b}_{u}),  
\end{aligned}
\end{equation}
where $\mathbf{W}_{*}$ and $\mathbf{U}_{*}$ are the respective weight matrices, $\mathbf{x}_{j}$ is the input at time step $j$, $\mathbf{h}_{j-1}$ is the previous hidden state at time step $j-1$, and $\mathbf{b}_{*}$ is the bias, where the subscript $*$ refers to the reset gate $r$, or the update gate $u$. In this scheme, the reset gate controls how much information will be forgotten from the previous hidden state, and the update gate controls the flow of information allowed to pass from the previous hidden state to the next hidden state \cite{chung2014empiricalevaluationgatedrecurrent}. 

Then a candidate hidden state, $\mathbf{\tilde{h}}_{j}$, is computed,
\begin{equation}
    \mathbf{\tilde{h}}_{j} = \tanh(\mathbf{W}_{h} \cdot \mathbf{x}_{j} + \mathbf{U}_{h}(\mathbf{r}_{j} \odot \mathbf{h}_{j-1} ) + \mathbf{b}_{h} )
\end{equation}
Lastly, the hidden state, $\mathbf{h}_{j}$ is computed,
\begin{equation}
    \mathbf{h}_{j} = \mathbf{u}_{j} \odot \mathbf{h}_{j-1} + (1-\mathbf{u}_{j}) \odot \mathbf{\tilde{h}}_{j}
\end{equation}
The gating structure in GRUs was heavily inspired by that of LSTMs. GRUs offer a more compact representation while still offering a notion of cell memory, enabling long-term memory.

RNNs, LSTMs, and GRUs have established the core mechanisms by which neural networks can capture temporal dependencies, mitigate vanishing gradients, and model long-range patterns in sequential data \cite{LSTM}. However, in real-world applications, these models face challenges such as interpretability and accuracy \cite{bengio1994learning}. To address these challenges, many researchers have developed novel architectures to extend these models.

Attention-augmented gated recurrent unit (GRU) \cite{NIU20231,yu2019attention} is a variant of GRU with an attention mechanism that learns a weighted summary of the GRU hidden states, allowing the classifier to focus on the most informative time steps. Specifically, given the hidden state sequence $\{h_1, h_2, \ldots, h_T\}$ produced by the GRU, the attention layer computes attention scores as $e_j = \mathbf{v}^{\top} \tanh(\mathbf{W} \mathbf{h}_j),$ which are normalized by a softmax function $\alpha_j = \mathrm{softmax}(e_j),$ to obtain attention weights $\alpha_j$ for $j = 1 \ldots T$.  The context vector is then calculated as $\mathbf{c} = \sum_{j=1}^{T} \alpha_j \mathbf{h}_j$. This context vector $c$ is concatenated with or used to replace the final GRU hidden state and passed to the output layer for classification.

\textbf{Applications:} The model was used in \cite{zhang2023radar} for radar-specific emitter identification (SEI). It was trained on denoised unintentional phase-modulation-on-pulse (UPMOP) sequences extracted from radar pulse streams, which serve as robust fingerprints under missing or spurious pulses and low signal-to-noise ratio (SNR) conditions. These data were derived from electronic support and electronic intelligence (ELINT) operational scenarios. Experimental results showed that the proposed attention-GRU model achieved over 93\% accuracy in identifying individual radar emitters under low-SNR conditions, outperforming non-attention GRU baselines.

\subsection{Transformers}\label{sec:transformers}
To extend the capacity of convolutional and recurrent networks, many sequence tasks, such as machine translation and abstractive summarization, require modeling long-range dependencies, where RNNs face vanishing-gradient limits and CNNs need very large receptive fields \cite{bengio1994learning,bai2018empirical}. Transformers, first introduced by Vaswani et al. \cite{vaswani2017attention}, replace recurrence and convolution with self-attention and set a new state of the art on translation. A transformer is a neural network architecture that uses self-attention to compute context-dependent representations of sequences. Unlike recurrent or convolutional networks, transformers process tokens in parallel and model pairwise interactions with attention weights.

An input $\mathbf{X}\in\mathbb{R}^{n\times d}$ contains $n$ token embeddings, one row per token with feature width $d$. Linear projections map each token to queries, keys, and values using $\mathbf{W}_Q,\mathbf{W}_K\in\mathbb{R}^{d\times d_k}$ and $\mathbf{W}_V\in\mathbb{R}^{d\times d_v}$, producing
$\mathbf{Q}=\mathbf{X}\mathbf{W}_Q,\quad \mathbf{K}=\mathbf{X}\mathbf{W}_K,\quad \mathbf{V}=\mathbf{X}\mathbf{W}_V.$ The scaled score matrix $\mathbf{Q}\mathbf{K}^\top/\sqrt{d_k}\in\mathbb{R}^{n\times n}$ contains dot-product similarity scores for every query–key token pair. A mask $\mathbf{M}\in\mathbb{R}^{n\times n}$ encodes constraints such as causal blocking by placing large negative entries where attention is disallowed and zeros elsewhere. Row-wise normalization with $\mathrm{softmax}$ gives attention weights that sum to one across keys and produce the contextualized representation

\begin{equation}
\mathrm{Attn}(\mathbf{Q},\mathbf{K},\mathbf{V})=\mathrm{softmax}\!\left(\frac{\mathbf{Q}\mathbf{K}^\top}{\sqrt{d_k}}+\mathbf{M}\right)\mathbf{V}\in\mathbb{R}^{n\times d_v}.
\end{equation}

Softmax converts any real-valued score vector into a probability vector by exponentiating the entries and normalizing so they are positive and sum to one, which amplifies larger scores while keeping a smooth sensitivity to all inputs.
Multi-head attention (MHA) runs this computation over $h$ heads by splitting the channel dimensions, concatenates the $h$ outputs, and projects back to the model width with $\mathbf{W}_O\in\mathbb{R}^{h d_v\times d}$. The sublayer output adds a residual connection. In the common pre-norm arrangement, layer normalization ($\mathrm{LN}$) is applied before the MHA function, resulting in the output

\begin{equation}
\mathbf{H}=\mathrm{MHA}(\mathrm{LN}(\mathbf{X}))+\mathbf{X},
\end{equation}

which preserves information and stabilizes gradients. Layer normalization operates per token across its $d$ features. For a token vector $\mathbf{x}\in\mathbb{R}^d$, it computes the mean $\mu(\mathbf{x})$ and standard deviation $\sigma(\mathbf{x})$ over the $d$ coordinates, rescales the centered vector $(\mathbf{x}-\mu(\mathbf{x}))/\sigma(\mathbf{x})$, then applies a learned transform with scale and shift parameters $\boldsymbol{\gamma},\boldsymbol{\beta}\in\mathbb{R}^d$, that recover any needed scale and offset, which yields $\mathrm{LN}(\mathbf{x})=\boldsymbol{\gamma}\odot\big((\mathbf{x}-\mu(\mathbf{x}))/\sigma(\mathbf{x})\big)+\boldsymbol{\beta}$. The same transform is applied independently to each token in the sequence.
A position-wise MLP then transforms each token independently, using two learned linear maps with nonlinearity $\sigma$ in between. For $\mathbf{h}\in\mathbb{R}^d$, let the expansion factor be $\alpha>1$, the hidden vector is $\mathbf{z}=\sigma(\mathbf{W}_1\mathbf{h}+\mathbf{b}_1)$,  with $\mathbf{W}_1 \in \mathbb{R}^{(\alpha d)\times d}, \mathbf{b}_1 \in \mathbb{R}^{\alpha d}$; followed by the projection back to model width $\mathrm{MLP}(\mathbf{h})=\mathbf{W}_2{\mathbf{z}}+\mathbf{b}_2$, with $\mathbf{W}_2 \in \mathbb{R}^{d \times(\alpha d)}, \mathbf{b}_2 \in \mathbb{R}^{d}$. With pre-norm and a residual connection, the layer output is $\mathbf{Y}=\mathrm{MLP}(\mathrm{LN}(\mathbf{H}))+\mathbf{H},$ so tensors at the layer boundary retain shape $n\times d$.


\textbf{Applications:} Transformers have recently been used used in image classification, detection, generation, cross-modal retrieval and zero-shot recognition, audio tagging, protein structure prediction, long-horizon time-series forecasting \cite{dosovitskiy2021an,Radford2021LearningTV,gong2021astaudiospectrogramtransformer,Jumper2021highly,Zhou2020InformerBE,chen2021decision}. Although transformers are central, graph neural networks continue to evolve and often integrate attention, from Graph Attention Networks \cite{velickovic2018graph} to transformer-style graph models such as Graphormer that report strong graph-level results and competitive leaderboard performance \cite{ying2021transformers}.

\subsection{Graph Neural Networks}
At the core of modern military operations lies a web of intricate relationships among personnel, systems, information, and infrastructure. These inherently relational dynamics make graph-based data representations particularly powerful, enabling analysts to observe connections in real time and make more informed decisions \cite{snyder2024aprimer}. Incorporating both node features and network structure can yield more powerful predictions making relational information critical \cite{hamilton2017inductive}. To rigorously capture such relationships, graph theory provides a formal framework for modeling entities and their interactions.

Formally, a graph, $\mathcal{G} =(\mathcal{V}, \mathcal{E})$, is defined as a set of vertices, $\mathcal{V}$, also called nodes or points, and a set of edges, $\mathcal{E}$. The edges describe the relationship between two nodes, that is, for $u,v \in \mathcal{V}$, $\exists(u,v) \in \mathcal{E}$. In practice, a node could represent an agent that is present in a scenario, and an edge can represent the distance from one agent to another if they are related. 

For computational purposes, it is convenient to encode the graph structure algebraically. An adjacency matrix, $\mathbf{A} \in \mathbb{R}^{|\mathcal{V}| \times |\mathcal{V}|}$, is a sparse matrix commonly used to represent a graph. To create an adjacency matrix, the nodes in the graph must be ordered, $v_{i} \in \mathcal{V}$, $i=1, \dots , |\mathcal{V}|$, so that each row and column of $\mathbf{A}$ corresponds to a particular node. For an unweighted graph, $[\mathbf{A}]_{i,j} = 1$ if $(v_{i} , v_{j}) \in \mathcal{E}$, otherwise $[\mathbf{A}]_{i,j} = 0$. For a weighted graph, the adjacency matrix may contain values other than 0 and 1. These weighted edges may represent the strength of the relationship or the distance between two nodes. For an undirected graph, $\mathbf{A}$ will be symmetric. This is not necessarily true for a directed graph, as certain nodes may have more influence on the relationship between two nodes. 

A degree matrix, $\mathbf{D} \in \mathbb{R}^{|\mathcal{V}| \times |\mathcal{V}|}$, is used to summarize node connectivity. The degree of node $v_{i} \in \mathcal{V}$ is defined as $d_{i} = \sum_{j \in |\mathcal{V}|} \mathbf{A}_{i,j}$. The degree matrix, $\mathbf{D}$, is defined as $\mathbf{D} = diag(d_{1}, \dots , d_{|\mathcal{V}|})$. The notion of degree can vary for weighted, unweighted, directed, and undirected graphs. 

A particularly important construction that arises from $\mathbf{A}$ and $\mathbf{D}$ is the graph Laplacian, defined as $\mathbf{L} = \mathbf{D} - \mathbf{A}\in \mathbb{R}^{|\mathcal{V}| \times |\mathcal{V}|}$. This version of the graph Laplacian is often referred to as the unnormalized graph Laplacian. There are two popular ways to normalize a graph Laplacian. The symmetric normalized graph Laplacian is defined as $\mathbf{L}_{sym} = \mathbf{D}^{-\frac{1}{2}}\mathbf{A} \mathbf{D}^{-\frac{1}{2}} \in \mathbb{R}^{|\mathcal{V}| \times |\mathcal{V}|}.$ The random walk normalized graph Laplacian is defined as $\mathbf{L}_{rw} = \mathbf{D}^{-1}\mathbf{L} =\mathbf{D}^{-1} (\mathbf{D} - \mathbf{A}) \in \mathbb{R}^{|\mathcal{V}| \times |\mathcal{V}|}.$ The Laplacian encodes both connectivity and degree information, and has fundamental properties that make it central to spectral graph theory. The various normalized forms of the graph Laplacian can be especially useful for different tasks like spectral clustering or partitioning \cite{vonluxburg2007tutorialspectralclustering}.

In a general graph neural network scheme, some form of message passing is done by vector ``messages'' passing from one node to the next. Graph neural networks (GNNs) take the input graph $\mathcal{G} = (\mathcal{V}, \mathcal{E})$ with a set of node features, $\mathbf{X} \in \mathbb{R}^{d \times | \mathcal{V}|}$ to create node embeddings $\mathbf{z}_{u},$ for $u \in \mathcal{V}$. At each layer $l$ of a GNN, a hidden embedding is created, $\mathbf{h}_{u}^{(l)}$, for each node $u \in \mathcal{V}$. First, a ``message'' is gathered or aggregated 

\begin{equation}\label{eq:gnn_message}
    \mathbf{m}_{\mathcal{N}_{u}}^{(l)} = f^{(l)}( \{\mathbf{h}_{v}^{(l)}: v \in \mathcal{N}_{u}\}), 
\end{equation}

where $\mathcal{N}_{u}$ is the neighborhood of some node $u \in \mathcal{V}$ and $f^{(l)}$ is some differentiable function that aggregates the neighborhood information. Since $f$ is defined over a set of hidden embeddings, $f$ must be permutation equivariant to handle this lack of ordering. Then, the new hidden embedding is created by updating the previous hidden embedding 
\begin{equation}\label{eq:gnn_embed}
     \mathbf{h}_{v}^{(l+1)} = \Upsilon^{(l)} (\mathbf{h}_{v}^{(l)}, \mathbf{m}_{\mathcal{N}_{u}}^{(l)}),
\end{equation}

where $\Upsilon^{(l)}$ is some differentiable function. The initial hidden embedding at $k=0$ is usually set to be the input feature vector of the respective node, that is, $\mathbf{h}^{0}_{v}  = \mathbf{x}_{v},$ for $ v \in 
\mathcal{V}$. The node embedding, $\mathbf{z}_{u},$ for $ u \in \mathcal{V}$, is given by the output of the final layer message-passing layer of the GNN, $\mathbf{z}_{u}  = \mathbf{h}_{u}^{*},$ for $u \in \mathcal{V}$.

The original GNN was presented in \cite{MerkwirthGNN} and \cite{OriginalGNN}. In these works, hidden embeddings are defined as follows 
\begin{equation}
    \mathbf{h}_{u}^{(l)} = \sigma \Big( \mathbf{W}_{self}^{(l)} \mathbf{h}_{u}^{(l-1)}  + \mathbf{W}_{neigh}^{(l)} \sum_{v \in \mathcal{V}} \mathbf{h}_{v}^{(l-1)}\Big),
    \label{eq:BasicGNN}
\end{equation}

where $\mathbf{W}_{self}^{(l)}, \mathbf{W}_{neigh}^{(l)} \in \mathbb{R}^{d^{(l)} \times d^{(l-1)}}$ are trainable weight matrices and $\sigma$ is a non-linear activation.

Equation \ref{eq:BasicGNN} is given at the node-level definition. This can be extended to a graph-level definition as follows 

\begin{equation}
    \mathbf{H}^{(l)} = \sigma \Big(  \mathbf{H}^{(l-1)} \mathbf{W}_{self}^{(l)} + \mathbf{A}  \mathbf{H}^{(l-1)} \mathbf{W}_{neigh}^{(l)}\Big),
\end{equation}

 where $\mathbf{H}^{(l)} \in \mathbb{R}^{| \mathcal{V}| \times d}$ is the matrix of node embeddings at layer $l$ and $\mathbf{A} $ is the adjacency matrix.  

Given the growing availability and operational importance of graph-structured data, a large body of work has developed GNNs that learn over connectivity, including spectral or convolutional methods such as Graph Convolutional Networks, spatial message-passing methods such as Graph Sample and Aggregate (GraphSAGE), attention-based architectures such as Graph Attention Networks, and temporal variants designed for dynamic graphs.

\subsubsection{Graph Convolutional Networks}
Spectral-based GNNs define convolutions in the spectral domain, originally through the eigendecomposition of the graph Laplacian \cite{bruna2014spectral,zhou2021graphneuralnetworksreview}. However, recent methods approximate these operations and avoid explicit eigendecomposition for efficiency and scalability \cite{defferrard2016convolutional,kipf2017semi}.  A classic example is the graph convolutional network (GCN) \cite{kipf2017semi}, which smooths the features of the nodes over the graph by applying a normalized adjacency operator.  Given a graph $\mathcal{G} = (\mathcal{V}, \mathcal{E})$ with vertices $\mathcal{V}$ and edges $\mathcal{E}$, the layer update rule is

\begin{equation}\label{eq:GCN_update}
    \mathbf{H}^{(l+1)} = \sigma\left(\hat{\mathbf{D}}^{-\tfrac{1}{2}} \hat{\mathbf{A}} \hat{\mathbf{D}}^{-\tfrac{1}{2}} \mathbf{H}^{(l)} \mathbf{W}^{(l)}\right),
\end{equation}

where $\mathbf{\hat{A}}=\mathbf{A}+\mathbf{I}$ is the adjacency matrix of the graph with self-loops added, $\mathbf{\hat{D}}$ is the corresponding diagonal degree matrix,  $\mathbf{H}^{(l)}$ contains the node feature embeddings at layer $l$, and $\mathbf{W}^{(l)} $ is a trainable weight matrix that transforms the features at level $l.$ The function $\sigma(\cdot)$ denotes a nonlinear activation function. Spectral GCNs aggregate neighbor information through adjacency and degree matrices, capturing local structure and revealing hidden subnetworks, anomalies, and hidden connections. 

\textbf{Applications:} Graph convolutional neural networks are effective across standard graph tasks \cite{kipf2017semi,kipf2016variational}, text applications that operate over dependency trees, including relation extraction and semantic role labeling \cite{zhang2018graph,marcheggiani2017encoding}, social-network analytics such as rumor and fake-news detection \cite{Monti2019FakeND}, recommendation on user–item bipartite graphs using stacked GCN layers \cite{wand2019neural}, neuroimaging on brain connectomes for subject-level classification \cite{PARISOT2018117}, and traffic and sensor networks where a static GCN backbone models spatial structure alongside temporal modules \cite{STGNN_Yu_2018}.
\subsubsection{GraphSAGE and GraphSAINT}

Spatial-based GNNs define convolution directly in the graph domain by aggregating information from local neighborhoods without relying on spectral theory \cite{hamilton2017inductive,xu2019how}. Instead of using the adjacency matrix for spectral filtering, they perform message passing using flexible aggregation functions. A common problem with spatial-based GNNs is the ``exploding neighbor'' problem \cite{NEURIPS2021_a378383b}. This issue occurs when using a spatial-based method on a large graph, as the number of neighbors to consider increases exponentially with each additional layer. To mitigate this problem, sampling methods like GraphSAGE \cite{hamilton2017inductive} and GraphSAINT \cite{zeng2020graphsaint} were introduced. 

GraphSAGE \cite{hamilton2017inductive} updates each node's hidden features $\mathbf{h}_u^{(l)}$ by sampling and aggregating over a fixed number of neighbor features:

\begin{equation}
\mathbf{h}_u^{(l+1)} = \sigma\left(\mathbf{W}^{(l)} \cdot \operatorname{AGG}\left(\left\{\mathbf{h}_u^{(l)}\right\} \cup \left\{\mathbf{h}_v^{(l)}, \, v \in \mathcal{N}(u)\right\}\right)\right),
\end{equation}

where AGG denotes aggregation and can indicate a mean, sum, or pooling function, and $\mathcal{N}_{u}$ is the neighborhood of some node $u \in \mathcal{V}$.

US Army researchers introduced GraphSAINT \cite{zeng2020graphsaint}, a sampling-based graph neural network technique. Unlike traditional GNN approaches that suffer from scalability and sampling bias when handling large graphs, GraphSAINT introduces stochastic subgraph sampling and applies loss and aggregation normalization to produce unbiased estimators of the full-batch gradient. Specifically, GraphSAINT samples subgraphs using strategies such as node, edge, or random walk sampling to obtain a sampled subgraph $\mathcal{G}_{s} = (\mathcal{V}_{s}, \mathcal{E}_{s})$, then constructs a GCN on $\mathcal{G}_{s}$. Although this sampling technique alleviates the ``neighbor explosion'' problem, it introduces bias into the estimation. To eliminate biases, GraphSAINT  uses the following hidden node embedding rule:

\begin{equation}
    \mathbf{h}_u^{(l+1)} = \sum_{v \in \mathcal{V}} (\alpha_{v,u})^{-1} \tilde{\mathbf{A}}_{u,v}\Big(\mathbf{W}^{(l)}\Big)^{T}  \mathbf{h}_{v}^{(l)} \mathds{1}_{v|u},
\end{equation}

where $\tilde{\mathbf{A}} = \mathbf{D}^{-1}\mathbf{A}$ is the normalized adjacency matrix, $\mathbf{A}$ is the adjacency matrix and $\mathbf{D}$ is the diagonal degree matrix, and  $\tilde{\mathbf{A}}_{u,v}$ is the element in the $(u,v)$ position in $\tilde{\mathbf{A}}$, $\alpha_{v,u}$ is the aggregator normalization constant, $\Big(\mathbf{W}^{(l)}\Big)^{T} \in \mathbb{R}^{d^{(l)} \times d^{(l-1)}}$ is a trainable weight matrix, and $\mathds{1}_{v|u}$ is an indicator function given $u$ is in the subgraph (i.e. $\mathds{1}_{v|u} = 0$ if $u \in \mathcal{V}_{s} \wedge (u,v) \notin \mathcal{E}_s$. This technique allows for efficient mini-batch training without compromising multi-hop message passing or structural fidelity. 

\textbf{Applications:} These models are the engine behind major ways companies suggest content and products. They help decide what items to show next to what you're viewing, how to rank new items, and quickly pull up relevant results on huge online platforms \cite{ying2018graph}. They also play a critical role in finding fraud and risk, such as checking financial transactions, spotting fake-identity rings, and detecting when accounts are stolen, even as new accounts pop up all the time \cite{motie2024financial}. Furthermore, they improve advertising and search results by using patterns of user clicks and interactions to better match queries with documents and find new potential customers \cite{wu2022graph}. Finally, they help analyze massive social and knowledge networks by training on smaller, representative parts of the whole network to learn accurately without having to process the entire huge graph at once \cite{zeng2020graphsaint}.
\subsubsection{Graph Attention Networks}

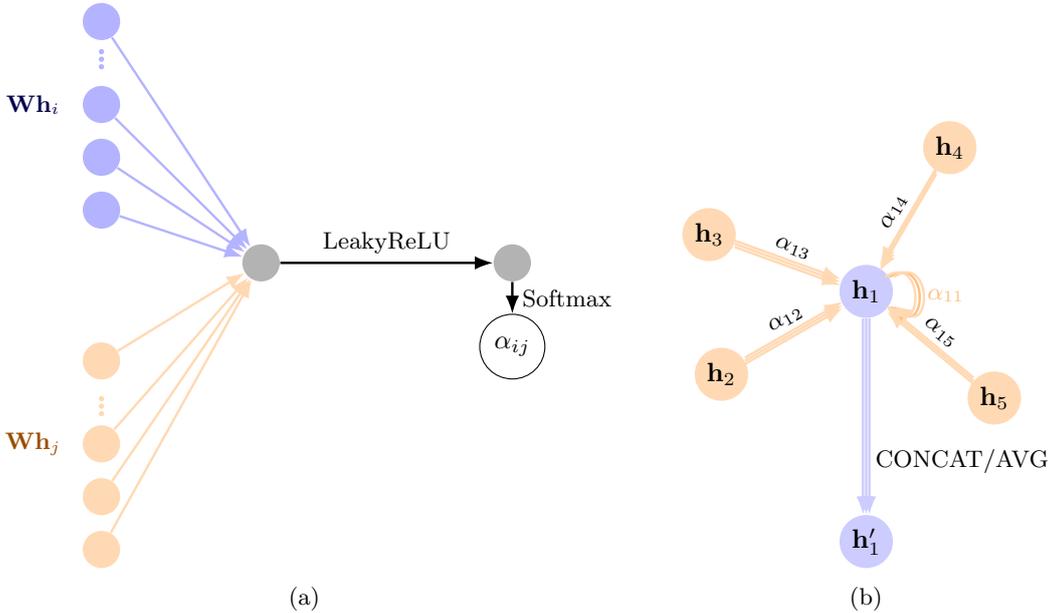
\begin{figure*}[!t]
    \centering
    \subfloat[]{\begin{tikzpicture}[x=1cm,y=1cm]

\tikzset{
  node/.style={circle,draw=none,minimum size=14pt,inner sep=0pt},
  blu/.style={node,fill=blue!30},
  ora/.style={node,fill=orange!30},
  dot/.style={circle,draw=none,minimum size=2pt,inner sep=0pt},
  blus/.style={dot,fill=blue!30},
  oras/.style={dot,fill=orange!30},
  grayn/.style={node,fill=gray!60},
  lab/.style={font=\small},
  arr/.style={-{Latex[length=2.4mm]},line width=.9pt},
}
\foreach \i/\y in {1/2.2,2/1.1,3/0.4,4/-0.3}{
  \node[blu] (bi\i) at (-4,\y) {};
}
\foreach \i/\y in {1/-4.8,2/-4.1,3/-3.4,4/-2.3}{
  \node[ora] (bj\i) at (-4,\y) {};
}

\foreach \y in {1.8,1.7,1.6}{
  \node[blus] at (-4,\y) {};
}
\foreach \y in {-3.0,-2.9,-2.8}{
  \node[oras] at (-4,\y) {};
}
\node (Wi) at (-4.9,1.1) [blue!30!black] {\small $\mathbf{W} \mathbf{h}_i$};
\node (Wj) at (-4.9,-3.4) [orange!60!black] {\small $\mathbf{W} \mathbf{h}_j$};

\node[grayn] (m1) at (-1.9,-1) {};
\node[grayn,right=2.8 of m1] (m2) {};
\draw[arr] (m1) -- node[above,lab]{LeakyReLU} (m2);
\node[draw,circle,minimum size=16pt,below=12pt of m2] (aij) {${\alpha}_{ij}$};
\draw[arr] (m2) -- node[right,lab]{Softmax} (aij);

\foreach \i in {1,2,3,4}{
  \draw[arr,blue!30] (bi\i) -- (m1);
  \draw[arr,orange!30] (bj\i) -- (m1);
}

\end{tikzpicture}}
    \label{fig:gata}
    \hfil
    \subfloat[]{\begin{tikzpicture}[x=1cm,y=1cm]

\tikzset{
  node/.style={circle,draw=none,minimum size=14pt,inner sep=0pt},
  blu/.style={node,fill=blue!20},
  ora/.style={node,fill=orange!30},
  dot/.style={circle,draw=none,minimum size=2pt,inner sep=0pt},
  blus/.style={dot,fill=blue!20},
  oras/.style={dot,fill=orange!30},
  grayn/.style={node,fill=gray!60},
  lab/.style={font=\small},
  arr/.style={-{Latex[length=2.4mm]},line width=.9pt},
}
\begin{scope}[xshift=9.2cm]
\node[blu,minimum size=20pt] (c) at (0,0) {$ \mathbf{h}_1$};

\foreach \ang/\name/\lab in {210/n2/2,160/n3/3,60/n4/4,-40/n5/5}{
  \node[ora,minimum size=20pt] (\name) at (\ang:2.2) {$ \mathbf{h}_{\lab}$};
  \foreach \s in {-1.2pt,0pt,1.2pt}{
    \draw[arr,orange!30,transform canvas={yshift=\s}] (\name) -- (c);
  }
}

\draw[arr,orange!50]
  (c) .. controls +(0.8,0.5) and +(0.8,-0.5) ..
  node[pos=.55,right,lab] {$\alpha_{11}$} (c);
\foreach \s in {-1.6pt,1.6pt}{
  \draw[arr,orange!30,transform canvas={xshift=\s}]
    (c) .. controls +(0.8,0.5) and +(0.8,-0.5) .. (c);
}

\path (n2) -- node[lab,sloped,above] {$\alpha_{12}$} (c);
\path (n3) -- node[lab,sloped,above] {$\alpha_{13}$} (c);
\path (n4) -- node[lab,sloped,above] {$\alpha_{14}$} (c);
\path (n5) -- node[lab,sloped,above] {$\alpha_{15}$} (c);

\node[blu,below=2.6cm of c,minimum size=20pt] (out) {$ \mathbf{h}'_1$};
\foreach \s in {-1.2pt,0pt,1.2pt}{
  \draw[arr,blue!20,transform canvas={xshift=\s}] (c) -- (out);
}
\node[lab,right] at ($(c)!0.68!(out)$) {CONCAT/AVG};
\end{scope}
\end{tikzpicture}}
    \label{fig:gatb}
    \caption{GAT methodology. (a) For a pair of neighboring nodes $i$ and $j$, node features are first linearly transformed ($\mathbf{W}\,{\mathbf{h}}_{i}$, $\mathbf{W}\,{\mathbf{h}}_{j}$), concatenated, passed through a LeakyReLU, and normalized with a softmax to yield attention coefficients ${\boldsymbol{\alpha}}_{ij}$.
  (b) Each node aggregates its neighbors’ features (including a self-loop) using the learned weights ${\boldsymbol{\alpha}}_{ij}$. Outputs from multiple heads are combined to produce the updated representation ${\mathbf{h}}^{\,\prime}$.}
  \label{fig:gat}
\end{figure*}

Inspired by the self-attention paradigm originally introduced in sequence modeling tasks, which later became foundational in transformer architectures and large language models (LLMs) \cite{vaswani2017attention}, attention-based GNNs generalize the attention mechanism on graphs. The graph attention network (GAT) \cite{velickovic2018graph} extends the spatial-based GNN approach by utilizing self-attention, which involves learning attention weights for each neighbor, allowing the model to prioritize more informative connections and eliminate noise. The hidden node embedding update rule is defined as

\begin{equation}
    \mathbf{h}_u^{(l+1)} = \sum_{v \in \mathcal{N}_{u}} \alpha_{v,u} \mathbf{W}^{(l)} \mathbf{h}_{v}^{(l)} ,
\end{equation}

\begin{equation}
    \alpha_{v,u} = \frac{\text{exp} \Big( \sigma \Big( \mathbf{a}^{T} [ \mathbf{W}^{(l)} \mathbf{h}_{u} \| \mathbf{W}^{(l)} \mathbf{h}_{v}] \Big)\Big)}{ \sum_{y \in \mathcal{N}_{u}} \text{exp} \Big( \sigma \Big( \mathbf{a}^{T} [ \mathbf{W}^{(l)} \mathbf{h}_{u} \| \mathbf{W}^{(l)} \mathbf{h}_{y}] \Big)\Big) }    
\end{equation}

where $\mathcal{N}_{u}$ is the neighborhood of some node $u \in \mathcal{V}$, $\alpha_{v,u}$ are the attention coefficients, $\sigma$ is an activation function (LeakyReLU), $\mathbf{W}^{(l)} \in \mathbb{R}^{d^{(l)} \times d^{(l-1)}}$ is a trainable weight matrix, $\mathbf{a} \in \mathbb{R}^{2 d^{(l)}}$ is the transpose of the weights parameterizing a single-layer feedforward neural network, and $\|$ denotes vector concatenation. See Figure \ref{fig:gat}. 


\textbf{Applications:} In \cite{liu2025radar}, the authors applied GAT to radar point clouds to classify targets with higher accuracy and lower computational cost. Radar works in darkness, smoke, and bad weather, and many sensors operate on limited size, weight, and power platforms. The attention mechanism addresses this challenge by processing the most informative returns and their local relationships, which reduces false alarms and missed detections. In their tests, the GAT beat non-attention baselines while using modest resources, enabling on-board use at the edge on vehicles, small unmanned aircraft, or mobile ground radars. Thus, it supports faster identification, tracking, and provides a more reliable operation in cluttered or contested environments.

\subsection{Bayesian Neural Networks}\label{sec:bnn}

Bayesian neural networks (BNNs) are an extension of conventional neural networks in which the model parameters, such as weights and biases, are treated as random variables with probability distributions rather than fixed point estimates. This probabilistic formulation allows BNNs to capture epistemic uncertainty in the model, offering principled ways to quantify confidence in predictions. The concept was first formalized in early Bayesian treatments of neural networks \cite{neal2012bayesian, mckay1992practicalbayesianframework}, which showed that such models provide a natural framework for regularization, generalization, and uncertainty estimation. While classical deep learning methods often produce highly confident yet unreliable predictions, BNNs explicitly represent uncertainty \cite{gal2016dropoutbayesianapproximationrepresenting, NEURIPS2020_322f6246}, making them particularly valuable in domains such as safety-critical decision-making. As such, BNNs form a cornerstone in the broader effort to bring probabilistic reasoning into deep learning. 

In BNNs, the goal is to infer the posterior distribution of the model's weights given the data. That is, 

\begin{equation}
P(\mathbf{W} | D) = \dfrac{P(D | \mathbf{W}) P(\mathbf{W})}{P(D)} = \dfrac{P(D | \mathbf{W}) P(\mathbf{W})}{\int P(D| \mathbf{W}) P(\mathbf{W}) d\mathbf{W}}
\end{equation}

However, computing the normalization constant, $\int P(D| \mathbf{W}) P(\mathbf{W}) d\mathbf{W}$, is often an intractable problem due to high dimensionality or lack of a closed form solution. To circumvent this issue, Bayesian inference algorithms, like Bayes-by-Backprop (BBB) \cite{blundell2015weightuncertaintyneuralnetworks}, are used to approximate the posterior. BBB \cite{blundell2015weightuncertaintyneuralnetworks} is an implementation of stochastic variational inference that uses a reparameterization trick. 

Variational inference \cite{cover1999elements} approximates an intractable true posterior distribution, $p(z| y)$, with a tractable variational distribution, $q_\phi(z)$, by solving an optimization problem. Given observations $y$ and latent variables $z$ with a specified joint distribution $p(y,z)=p(y| z)p(z)$, we choose a simple, often factorized, family $\{q_\phi(z)\}$ indexed by a set of parameters $\phi$. The goal is to select the member of this family that is closest to the true posterior, which is achieved by minimizing their Kullback--Leibler (KL) divergence $\phi^\star=\arg\min_{\phi}\,\mathrm{KL}\!\left(q_\phi(z)\,\|\,p(z| y)\right)$.

The KL divergence from a distribution $P$ to $Q$ is defined on a measurable space $\mathcal{X}$ as:
\begin{equation}
   \mathrm{KL}(P\,\|\,Q)=\int_{\mathcal X} p(x)\,\log\!\frac{p(x)}{q(x)}\,\mu(dx), 
\end{equation}
where $p$ and $q$ are the densities of $P$ and $Q$ with respect to a base measure $\mu$. To minimize $\mathrm{KL}(q_\phi\,\|\,p(\cdot| y))$, we use the Evidence Lower Bound (ELBO), $\mathcal{L}(\phi)$, which is tractable:

\begin{equation}
    \mathcal L(\phi)=\mathbb E_{q_\phi}\!\big[\log p(y,z)-\log q_\phi(z)\big],
\end{equation}

where  $\mathbb{E}_{q_\phi}[f(X)]$ the expectation of $f(X)$ with respect to $q_\phi$, that is,

\begin{equation}
    \mathbb{E}_{q_\phi}[f(X)] = \int_{\mathcal{X}} {q_\phi}(x) f(x) \, \mu(dx).
\end{equation}

The relationship between the ELBO, the marginal likelihood, $\log p(y)$, and the KL divergence is given by the decomposition $\log p(y)=\mathcal L(\phi)+\mathrm{KL}\!\left(q_\phi\,\|\,p(\cdot| y)\right).$
Since $\log p(y)$ is a fixed constant with respect to $\phi$, maximizing the ELBO is equivalent to minimizing the KL divergence. Thus, increasing $\mathcal{L}(\phi)$ tightens the bound and improves the approximation of the true posterior. In practice, the variational distribution $q_\phi$ is chosen to be differentiable with respect to its parameters $\mathbf{ \phi}$ so that $\mathcal L(\phi)$ can be efficiently optimized. 

BBB uses a variational posterior $ q_{\phi}(\mathbf{\theta}) $ to approximate the true posterior $ p(\mathbf{\theta} 
|D) $. To learn $\phi$, introduce auxiliary noise $ \mathbf{\epsilon}\sim p(\mathbf{\epsilon}) $ (e.g., $ \mathcal{N}(\mathbf{0},\mathbf{I}) $) and sample weights via a deterministic transform $\mathbf{\theta} = g_{\phi}(\mathbf{\epsilon})$ (e.g.,  $\mathbf{\mu}_{\phi} + \mathbf{\sigma}_{\phi}\odot \mathbf{\epsilon}$).

The Kingma--Welling reparameterization trick~\cite{kingma2013auto} converts sampling from the variational posterior into a deterministic, differentiable transform of noise that does not depend on $\phi$. This lets gradients flow through the stochastic node:
\begin{equation}
\nabla_{\phi}\,\mathbb{E}_{q_{\phi}(\mathbf{\theta})}[f(\mathbf{\theta})]
= \nabla_{\phi}\,\mathbb{E}_{\mathbf{\epsilon}\sim p(\mathbf{\epsilon})}\!\left[f\!\left(g_{\phi}(\mathbf{\epsilon})\right)\right]
= \mathbb{E}_{\mathbf{\epsilon}\sim p(\mathbf{\epsilon})}\!\left[\nabla_{\phi} f\!\left(g_{\phi}(\mathbf{\epsilon})\right)\right].
\end{equation}

For example, let $q_{\phi}(\mathbf{\theta})$ be a Gaussian variational distribution parameterized by $\phi = (\mu, \nu)$, where $\nu = \text{log}(1+ \text{exp}(\rho))$ to ensure the variance is non-negative. Let $\mathbf{h}^{(l-1)}$ be the vector input, $\mathbf{W}^{(l)}$ be the weights, $\mathbf{b}^{(l)}$ be the biases and of a fully-connected layer. Let $\mu^{(l)}_{ij}$ ad $\rho^{(l)}_{ij}$ be variational parameters for each weight $\mathbf{W}^{(l)}_{ij}$ and let $\mu^{(l)}_{b,k}$ ad $\rho^{(l)}_{b,k}$ be variational parameters for each bias $[\mathbf{b}^{(l)}]_{k}$.
For a fully-connected layer $l$, we sample as follows
\begin{equation}
\begin{aligned}
    \epsilon^{(l)}_{ij} \sim \mathcal{N}(0,1), &\text{     } \mathbf{W}^{(l)}_{ij} = \mu^{(l)}_{ij} + \nu^{(l)}_{ij} \epsilon^{(l)}_{ij},\\
    \epsilon^{(l)}_{b,k} \sim \mathcal{N}(0,1), &\text{     } \mathbf{b}^{(l)}_{k} = \mu^{(l)}_{b,k} + \nu^{(l)}_{b,k} \epsilon^{(l)}_{b,k}
\end{aligned}
\end{equation}
The forward pass of layer $l$ is defined by 
$\mathbf{h}^{(l)} = \sigma(\mathbf{W}^{(l)}\mathbf{h}^{(l-1)} + \mathbf{b}^{(l)})$, where $\sigma(\cdot)$ is a nonlinear activation function.

\textbf{Applications:} Bayes-by-Backprop is used to add calibrated uncertainty to classifiers and regressors for risk-aware decisions in vision and medical imaging, where weight posteriors flag low-confidence cases for human review. It has been applied to CNNs and clinical segmentation or classification settings to cope with scarce and noisy labels, drive exploration in reinforcement learning via posterior sampling over value or policy networks, improving the exploration-exploitation trade-off \cite{blundell2015weightuncertaintyneuralnetworks}. BBB provides inductive uncertainty in sequence models (Bayesian RNN/LSTM) for forecasting, reliability, and remaining-useful-life prognostics, where temporal predictions need epistemic estimates \cite{fortunato2019bayesianrecurrentneuralnetworks}. It supports active learning and out-of-distribution detection by exposing prediction variance and mutual information from weight uncertainty, improving sample selection and safety filters relative to point-estimate nets \cite{jospin2022deep}.

    \section{Topological Methods}

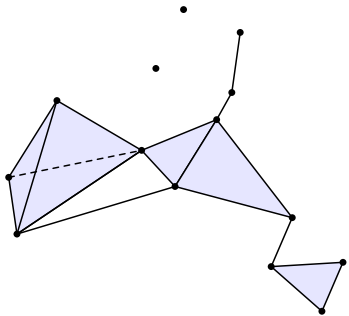
\begin{figure}[ht]
    \centering
    \resizebox{0.3\linewidth}{!}{\begin{tikzpicture}[line join=round,line cap=round,rotate=\ORIENT]
\tikzset{
  pt/.style={circle,fill=black,inner sep=1.3pt},
  ed/.style={black,thick},
  poly/.style={fill=blue!10,draw=black,thick},
  thinpoly/.style={fill=blue!10,draw=black,thick},
  lab/.style={font=\footnotesize,label distance=2pt}
}


\coordinate (M1) at (-1.3,1.7);
\coordinate (M2) at (0.2,2.5);
\coordinate (M3) at (1.1,0.9);
\coordinate (M4) at (-1.7,0.7);

\coordinate (R1) at (2.1,2.1);
\coordinate (R2) at (2.6,0.7);
\coordinate (R3) at (1.3,0.0);
\coordinate (R4) at (2.9,-1.6);

\coordinate (T1) at (2.1,-2.2);
\coordinate (T2) at (2.5,-3.4);
\coordinate (T3) at (3.3,-2.8);

\coordinate (C1) at (3.1,2.8);
\coordinate (C2) at (3.8,1.9);
\coordinate (C3) at (3.1,1.0);

\coordinate (P1) at (-3.6,3.2);
\coordinate (P2) at (-2.2,3.4);
\coordinate (P3) at (-4.6,-0.4);

\filldraw[poly] (M4) -- (M1) -- (M2) -- (M3) -- cycle;
\filldraw[poly] (M3) -- (R3) -- (R2) -- cycle;
\filldraw[poly] (R3) -- (R2) -- (R4) -- cycle;
\filldraw[poly] (T1) -- (T2) -- (T3) -- cycle;

\draw[ed] (M4)--(R3);
\draw[ed] (M4)--(M3);
\draw[ed] (M2)--(M4);
\draw[ed] (R2)--(C3);
\draw[ed] (C2)--(C3);
\draw[ed] (R4)--(T1);

\draw[dashed,thick] (M1) -- (M3);

\foreach \p in {M1,M2,M3,M4,R1,R2,R3,R4,T1,T2,T3,C1,C2,C3}
  \node[pt] at (\p) {};

\end{tikzpicture}}
    \caption{Simplicial complex} \label{fig:simplicial}
\end{figure}

A standard way to model the shape of data is with simplicial complexes built from basic building blocks called simplices. A $k-$simplex $\sigma_k =[v_0,\dots,v_k]$ is a $k$-dimensional shape obtained from $k+1$ vertices. For example, a 0-simplex is a point, a 1-simplex is a line segment, a 2-simplex is a filled-in triangle, and a 3-simplex is a solid tetrahedron. A simplicial complex is a collection of simplices glued together such that every face of a simplex is also included, and the intersection of any two simplices is either empty or a shared face \cite{MR1867354}. See Figure \ref{fig:simplicial}.

With this representation in place, persistent homology encodes structures such as connected components, loops, and voids into persistence diagrams (PDs) or barcodes to produce compact and noise-resistant representations of complex datasets. Small perturbations in the data produce only small changes in the diagram, so the summaries are stable. By capturing structure from local to global scales, PDs offer a comprehensive view of the organization in the signal \cite{cohen2007stability}. 

\begin{figure}
\centering
\begin{subfigure}[t]{0.30\linewidth}
  \centering
  \includegraphics[width=0.8\linewidth]{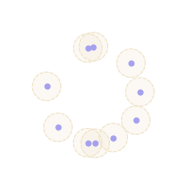}
  \subcaption{Union of balls at radius 0.3}
\end{subfigure}\hspace{0.1\linewidth}
\begin{subfigure}[t]{0.30\linewidth}
  \centering
  \includegraphics[width=0.8\linewidth]{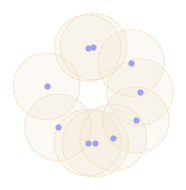}
  \subcaption{Union of balls at radius 0.7}
\end{subfigure}

\medskip
\begin{subfigure}[t]{0.45\linewidth}
  \includegraphics[width=\linewidth]{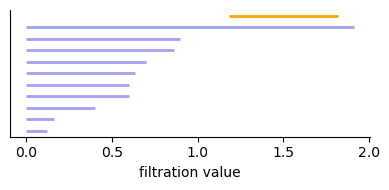}
  \subcaption{Persistence Barcode} 
\end{subfigure}\hfill
\begin{subfigure}[t]{0.45\linewidth}
  \includegraphics[width=\linewidth]{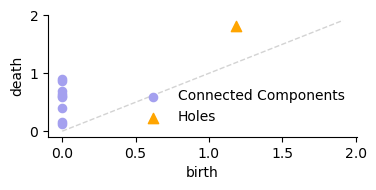}
  \subcaption{Persistence Diagram} 
\end{subfigure}
\caption{Persistent homology of a circular point cloud.}
\label{fig:persistence}
\end{figure}

To represent topological features in persistence diagrams and persistence barcodes, analysts apply a filtration of complexes, like the Vietoris–Rips filtration, by gradually increasing a distance threshold among data points, as shown in Figure \ref{fig:persistence}. At first, each point is isolated, but as the threshold grows, edges connect nearby points, forming connected components. This threshold, often denoted by a parameter $\epsilon$ or $r$, determines which points are considered “close enough” to be connected. When the pairwise distance between two points is less than or equal to this threshold, an edge is added between them. As $r$ increases, new simplices are formed whenever all pairwise distances between their vertices fall below the current threshold.. Larger thresholds fill in triangles and higher simplices, creating loops and higher-dimensional voids, which eventually disappear as the complexes become denser.

A PD is a collection of points plotted in the plane, where each point $(b,d)$ represents a topological feature of the data. The first coordinate shows the scale at which the feature first appears (birth), and the second coordinate marks the scale at which it disappears (death). The features that persist over a long range of scales correspond to points farther away from the diagonal $b=d$. These long-lived features are typically the most meaningful and are the primary focus of analysis, whereas points clustered near the diagonal usually represent short-lived features that are often treated as noise. Similarly, persistence barcodes represent each feature as a horizontal line segment whose endpoints mark its birth and death scales; long bars highlight significant, persistent structures, while short bars typically correspond to noise. For more information on PDs and filtration, see \cite{Carlsson2021tda}.

\subsection{Learning Frameworks in Persistence Diagram Space}

Integrating persistent homology with machine learning pipelines in classification, anomaly detection, and predictive modeling is highly relevant to military contexts where robustness and interpretability are critical \cite{Davies2022ARO}. However, PDs present a challenge in machine learning because their homological features lack intrinsic order, unlike the coordinates in vectors. This lack of structure makes it difficult to directly apply standard machine learning algorithms. To address this issue, researchers have developed a variety of techniques to map PDs into structured spaces that are more compatible with statistical and machine learning methods. Examples include vectorization approaches \cite{bubenik2015statistical,adams2017persistence}, and kernel methods tailored to PDs \cite{reininghaus2014asm,carrriere2017sliced}. These representations enable the use of algorithms such as support vector machines (SVMs), random forests, or neural networks on topological data.

\subsubsection{Clustering in Persistence Diagram Space}
Reducing PDs to a few summary statistics before being used in machine learning tasks removes important structural information, such as joint birth-death geometry, multimodal clusters in diagram space, long-versus short-persistence mixtures, and lifetime gaps. To address this, the authors of \cite{maroulas2017kmeans} extend the classical K-means clustering algorithm to the space of persistence diagrams by using the Wasserstein distance as a dissimilarity measure and the Fr\'{e}chet mean as the notion of cluster centroids. Their method groups time series by their underlying shape in phase space rather than by raw samples. The time series $x(t)$ form a point cloud by delay embedding. Each point is $z_t=(x(t),x(t+\tau),\ldots,x(t+(d-1)\tau))\in\mathbb{R}^d$, where $\tau$ is the delay and $d$ is the embedding dimension. Using the Euclidean metric\textbf{ $\rho$} on $\mathbb{R}^d$, a Vietoris–Rips complex is built on the set $P=\{z_t\}_{t=1}^k$ for a scale parameter $\varepsilon$. 
The \textbf{$p$-}Wasserstein distance, $W_p(D_1,D_2)=\left(\inf_{\gamma}\sum_{x\in D_1}\|x-\gamma(x)\|_\infty^{\,p}\right)^{1/p}$ , compares PDs $D_1$ and $D_2$, where $\gamma$ ranges over bijections between the multisets, that is, $\gamma : D_{1} \to D_{2}$, and matching to the diagonal is allowed to ensure a bijection exists. This distance measures the minimal total transport needed to align one diagram with the other under the max norm. The Wasserstein distance supplies a principled geometry on features.

The Fr\'{e}chet framework averages diagrams. Let $\mathcal{P}$ be the space of persistence diagrams with the Wasserstein metric and let $\mathcal{B}(\mathcal{P})$ be its Borel $\sigma$-algebra. For a probability measure $\mathbb{D}$ on $(\mathcal{P},\mathcal{B}(\mathcal{P}))$, define the Fr\'{e}chet functional $F^{W_p}_{\mathcal{P}}(D_1)=\int_{\mathcal{P}} W_p(D_1,D_2)^2\,d\mathbb{D}(D_2)$.
The Fr\'{e}chet variance is $\mathrm{Var}^{W_p}_{\mathbb{D}}=\inf_{D\in\mathcal{P}} F^{W_p}_{\mathcal{P}}(D)$, and the Fr\'{e}chet mean set is $\mathbb{E}^{W_p}(\mathbb{D})=\left\{D\in\mathcal{P}\;|\;F^{W_p}_{\mathcal{P}}(D)=\mathrm{Var}^{W_p}_{\mathbb{D}}\right\}$. For a finite dataset of diagrams, the empirical distribution admits Fr\'{e}chet means under standard conditions, so centroid updates are well defined. 

The clustering method developed on a non-Euclidean space starts with a dataset $\mathcal{D}=\{D_i\}_{i=1}^T$ of persistence diagrams and a suspected number of clusters $K$. Here, $K$ in $K$-means denotes the number of clusters, not the homology degree. By choosing $K$ initial centroids from $\mathcal{D}$, each diagram is assigned to the nearest centroid under \textbf{$W_p$}, and each centroid is replaced by a Fr\'{e}chet mean of the diagrams assigned to it. These steps are repeated until the cluster assignments stabilize. The result is a clustering of dynamics that respects topology, persists across noise, and admits a simple iterative algorithm based on assign and update under $W_p$.

Using synthetic datasets with signals that exhibit periodicity, bi-stability, chaos, and randomness, the proposed method is compared to dynamic time warping and wavelet-based clustering. Results show that the persistence-diagram-based approach achieves significantly lower error rates, especially in distinguishing periodic and bi-stable signals from random noise. The study highlights both the strengths and limitations of Wasserstein-based clustering, noting that while it excels at detecting large topological differences, it is less sensitive to finer geometric distinctions. 

Because Fr\'{e}chet means in PD space need not be unique, $K$-means steps can be unstable. This can be mitigated by clustering stable Euclidean embeddings of diagrams so centroids are unique in reproducing kernel Hilbert space (RKHS) \cite{kusano2016persistence}, or by using entropically regularized optimal transport Fr\'{e}chet means for robust centroid updates \cite{lacombe2018large}.

\subsubsection{Statistical Modeling of Random Persistence Diagrams}
To use statistics like inference, uncertainty quantification, hypothesis testing, and generative modeling on persistence diagrams, a probability distribution is needed. However, this is nontrivial because PDs are multisets without natural ordering and with variable size. The authors of \cite{maroulas2019nonparametric} define a global probability density for random persistence diagrams and construct a consistent kernel density estimator (KDE) that respects permutation invariance and variable cardinality. A random persistence diagram $D$ is a random finite multiset of points in $\mathbb{R}^2_{>}=\{(b,d):d>b\}$ with the Borel $\sigma$-algebra $\mathcal{B}(\mathbb{R}^2_{>})$. The belief function $\beta_D(S)=\mathbb{P}[D\subset S]$ for measurable $S\subset\mathbb{R}^2_{>}$ induces a global, permutation-invariant density $f_D$ on configurations through set derivatives. For $N$ points $\xi_1,\ldots,\xi_N$ with $\xi_i=(b_i,d_i)$, 
\begin{equation}
    \sum_{\pi\in S_N} f_D\!\big(\xi_{\pi(1)},\ldots,\xi_{\pi(N)}\big)
=\frac{\delta^N \beta_D}{\delta \xi_1\cdots \delta \xi_N}(\varnothing).
\end{equation}

Diagram cardinality yields a finite mixture representation $\beta_D(S)=a_0+\sum_{m=1}^{M} a_m\,q_m(S)$,  where $a_m=\mathbb{P}(|D|=m)$, $q_m(S)=\mathbb{P}\!\left[D\subset S\,\middle|\,|D|=m\right]$,  $a_0=\mathbb{P}(|D|=0)$, and $M$ is the maximal cardinality allowed by the model.

Given observed diagrams $D_1,\ldots,D_n$, the method defines a kernel $K_\sigma(Z,D)$, with bandwidth $\sigma>0$, that is a symmetric, permutation-invariant density on output diagrams $Z$ centered at $D$. The KDE of the global pdf is $\hat f(Z)=\frac{1}{n}\sum_{i=1}^n K_\sigma(Z, D_i)$.
Under standard regularity and a bandwidth schedule $\sigma\to 0$ at a suitable rate, $\hat f$ converges to the true global density in an appropriate sense.

The kernel separates features into two groups. Long-persistence features, which encode a stable topological structure, are modeled individually. Short-persistence features, which are numerous and often near the diagonal, are aggregated and modeled collectively as a ``cloud'' near the diagonal. This yields robustness to noise and improved computational efficiency while preserving prominent structure.

The probability hypothesis density (first-order intensity) $F_D(u)$ derived from the global density satisfies, for every measurable $U\subset\mathbb{R}^2_{>}$, $\mathbb{E}\big[\,|D\cap U|\,\big]=\int_U F_D(u)\,du,$ which gives the expected number of diagram points in $U$. The framework also introduces a dispersion statistic, the mean absolute bottleneck deviation (MAD), and establishes its convergence under the same asymptotic regime, providing a scalar measure of spread that complements the full density.

The construction supplies a rigorous statistical foundation for modeling randomness in persistence diagrams, enabling likelihood-based inference, hypothesis testing, and Bayesian methods. 

\textbf{Applications:} A key strength of the proposed nonparametric framework is its generality across data modalities. The method can be applied to any dataset that can be represented as a random persistence diagram, including time series, images, and higher-dimensional signals. The authors particularly emphasize its applicability to neuroscientific data, especially EEG (electroencephalographic) recordings.

In the context of EEG data, each time-series signal, recorded from multiple electrodes, can be treated as a high-dimensional trajectory that evolves over time. Instead of directly analyzing the signal, the topological features are extracted via persistent homology from time-delay embeddings of the EEG signal.These topological features encode complex dynamical patterns in brain activity-capturing, for example, the recurrence, synchronization, or oscillatory structures underlying cognitive processes.

The nonparametric density estimation developed in the paper then allows researchers to model the probability distribution of these random persistence diagrams across trials or subjects, quantify how the topological structure varies with the cognitive state or the type of stimulus, and perform hypothesis testing and classification based on topological statistics rather than raw signal amplitudes or frequencies. This is particularly valuable in neuroscience, where EEG data are noisy and highly nonstationary conditions under which traditional linear methods often fail to capture the underlying structure.

\subsubsection{Bayesian Modeling of Random Persistence Diagrams}
Although the global KDE offers a nonparametric, permutation-invariant density that is robust to noise and effective for exploratory or unsupervised learning tasks, it is less suited to class-conditional decisions or the explicit use of prior structural knowledge, which typically require additional modeling beyond the KDE \cite{GreenSeheult1988,Scott2015MDE,WandJones1994}. A recent advancement introduced a Bayesian statistical framework for persistence diagrams \cite{maroulas2020abayesian} that models diagrams as Poisson point processes (PPPs) on the tilted wedge $W=\{(b,\ell)\in\mathbb{R}^2:\ell>0\}$, with $d=b+\ell$, using conjugate Gaussian–mixture priors to obtain closed-form posterior intensities, then scoring and classifying diagrams via posterior diagram densities and Bayes factors.

\begin{figure}[ht]
    \centering
    \begin{subfigure}[t]{0.4\linewidth}
          \centering
          \includegraphics[width=\linewidth]{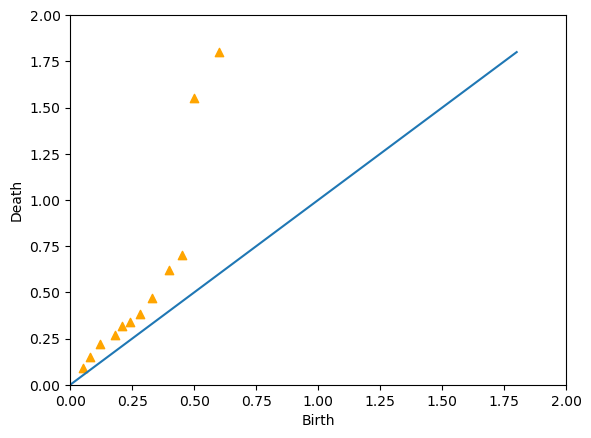}
          \subcaption{}
    \end{subfigure}\hspace{0.1\linewidth}
    \begin{subfigure}[t]{0.4\linewidth}
          \includegraphics[width=\linewidth]{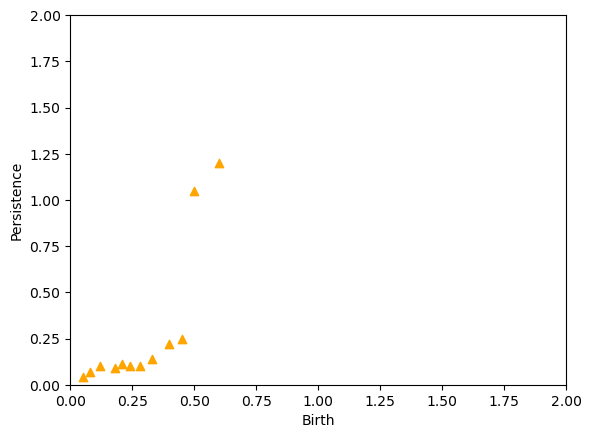}
          \subcaption{} 
    \end{subfigure}
    \caption{(a) A persistent diagram, (b) the corresponding tilted diagram.}
    \label{fig:tilted}
\end{figure}
A tilted diagram is treated as a finite collection of points modeled using a Poisson point process (PPP) with a spatially varying intensity function. For each class, training diagrams provide observations that inform the latent intensity of the corresponding class-level diagram. Observed points are assumed to arise from a combination of a matching process, which links latent points to observed ones, and a noise process that accounts for spurious or missing points. A weighting function balances these contributions, allowing the model to handle partial matching and dropout naturally. Choosing Gaussian mixtures for the latent intensity, substitution kernel, and noise ensures that posterior updates remain tractable and closed-form.

Given a set of training diagrams, the latent class intensity is updated to reflect both the prior structure and the observed points across all diagrams. This updated intensity can then be used to evaluate the likelihood of new diagrams under the class model. Classification between two classes is performed by comparing the relative likelihoods of the new diagram under each posterior intensity, producing a Bayes-factor-style decision metric that favors the class whose latent intensity better explains the observed features.

\textbf{Applications:} This framework allows for principled uncertainty quantification, hypothesis testing, and the incorporation of prior knowledge about topological features. The authors further demonstrate how posterior distributions over persistence diagrams can be used to construct Bayesian predictive densities, enabling likelihood-based inference. Several examples illustrate the methodology, including applications to noisy geometric data and dynamical systems. The formalization of persistence diagrams within Bayesian inference provides a coherent statistical machinery that opens up new avenues for TDA in scientific applications, ranging from shape recognition to complex system modeling.

\subsubsection{Bayesian Topological Signal Processing for Time-Series and Network Data}
Specializing this general framework to time series yields a concrete,
interpretable pipeline. \cite{oballe2022discrete} construct sublevel-set PDs from $s(t)$ and view them as tilted diagrams on $W$ by identifying persistence with lifetime ($p\equiv \ell$). The stochasticity of diagrams generated by random signals is described by the distribution obtained after applying the PD transform to the random signal. The Bayesian PPP model then gives a tractable approximation of that induced distribution via a posterior intensity learned from training diagrams.
\begin{figure}[ht]
  \centering
  \includegraphics[width=\textwidth]{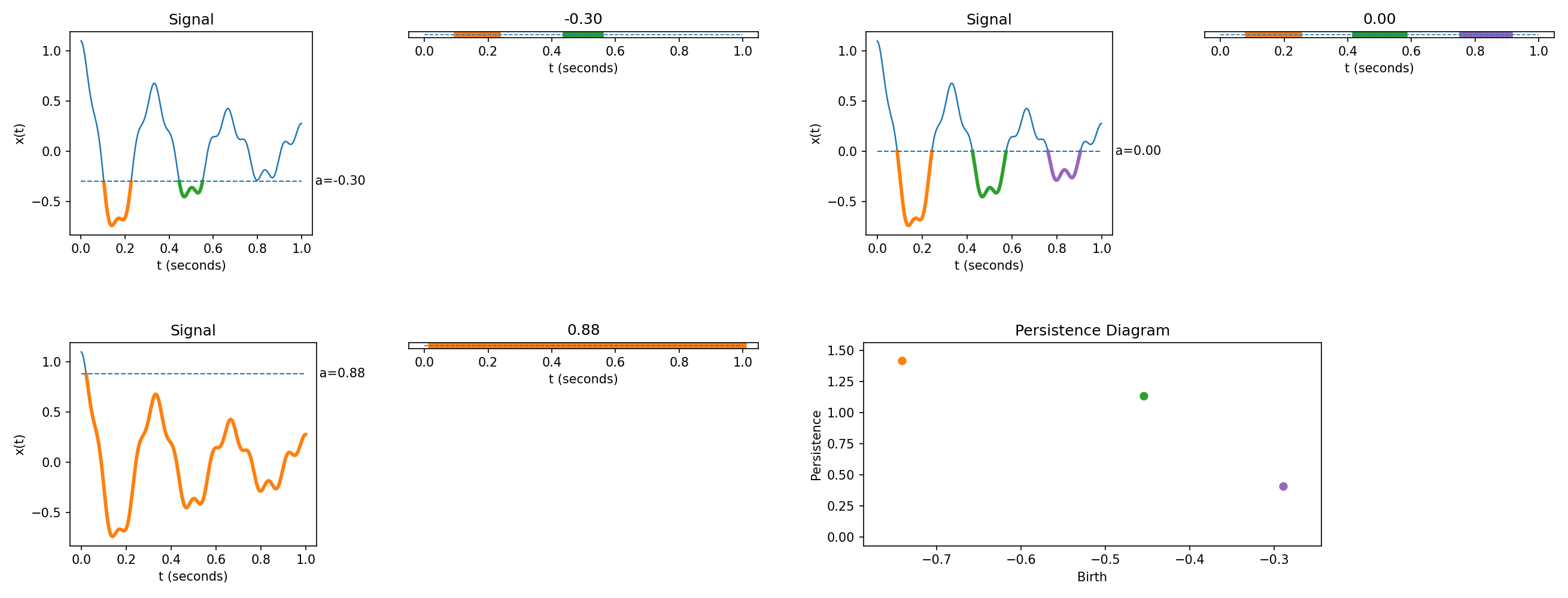}
  \caption{The sublevel sets of the damped signal at thresholds ${-0.30}$, ${0}$, and ${0.88}$ show how connected components of $\{t:\,x(t)\le a\}$ evolve with the level $a$. Colored segments denote distinct components, and the persistence diagram encodes the same filtration; point colors correspond to the components visible at $a=0$. As $a$ increases from $-0.30$ to $0$, short components appear and then attach to earlier ones. Raising $a$ further to $0.88$ produces merges where the later-born component is absorbed by the earlier-born component, so only longer-lived structures persist. The three points in the diagram record the associated birth values and lifetimes, matching the three colored components at $a=0$.
}
  \label{fig:damped_comp}
\end{figure}
In the signal setting this posterior intensity is a nonnegative intensity over birth–persistence whose magnitude indicates how often and how strongly features occur, and whose integral gives the expected number of features where dominant frequency governs diagram cardinality and the spread of birth locations, while instantaneous amplitude controls persistence magnitudes. Figure \ref{fig:damped_comp} illustrates these effects for a damped oscillation where stronger low-frequency content produces widely spaced minima and larger persistence, whereas increased high-frequency content compresses births and lowers extreme persistence. Thus the Bayesian update provides the decision-theoretic layer (posterior scoring and Bayes factors) on top of an interpretable topological summary that links classical spectral features to the geometry of persistence diagrams.

Bayesian topological signal processing (BTSP) has been applied across various domains such as brain connectivity by analyzing EEG and fMRI signals \cite{Gangapuram2024.05.02.592218}, detection of anomalies \cite{ramezani2021,xu2022scalable}, medical imaging and computer vision \cite{Veni2017ShapeCut}.

\textbf{Applications:} In \cite{maroulas2022bayesian}, the authors developed a Bayesian framework to classify the structure of biological networks, with a specific focus on actin filament networks in plant cells. The actin filament networks are converted into persistence diagrams, which highlight critical topological features that are directly relevant to cellular structure and function. Using Bayesian modeling, the framework quantifies variability in these features and incorporates prior biological knowledge, allowing for robust uncertainty estimation. The practical application of this methodology lies in its ability to classify and differentiate biological networks based on their structural patterns, providing valuable insights into cellular organization, cytoskeletal dynamics, and potentially informing experimental design in plant biology and related biomedical research.

\subsection{Topological Convolutional Neural Networks}\label{sec:tcnn}

\begin{figure}[ht]
  \centering
  \includegraphics[width=0.4\linewidth]{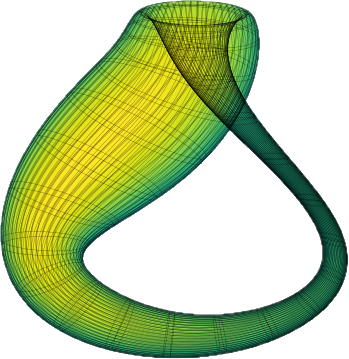}
  \caption{A 2-dimensional representation of the 3-dimensional immersion of the Klein bottle.}
  \label{fig:klein}
\end{figure}

In computer vision, convolutional neural networks (CNNs) have become the standard approach for extracting hierarchical features from images and videos \cite{zhao2024ai}. While CNNs, discussed in Section \ref{sec:CNN}, are effective for a wide range of image and video processing tasks, they do not explicitly exploit the underlying topological features of data. In contrast, topological convolutional neural networks (TCNNs) utilize manifolds such as the Klein bottle (see Fig. \ref{fig:klein}) and primary circle to parameterize image and video filters, and thereby enhance learning speed, data efficiency, generalizability, and interpretability compared to conventional CNNs \cite{love2020tcnn}. 

\begin{figure}[ht]
    \centering
    \includegraphics[width=0.5\textwidth]{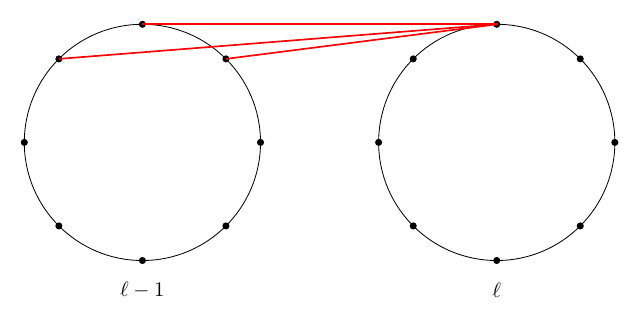} 
    \caption{A visual representation of the COL. The nodes on each circle represent the convolutional filters in layers $\ell-1$ and $\ell$ in a neural network. The red lines represent which filters in layer $\ell-1$ are connected to a particular filter in layer $\ell$ (i.e., the weights between the connected filters are nonzero).}
    \label{fig:circle one layer}
\end{figure}

TCNNs are constructed by replacing standard convolutional layers with topological convolutional layers, using predefined filters and localized connections based on topological metrics. There are two main types of topological convolutional layers: the circle filter (CF) and Klein filter (KF) layers, which use predefined filters based on embeddings of manifolds; and the circle one layer (COL) and Klein one layer (KOL), which prune connections based on topological distances (see Fig. \ref{fig:circle one layer} for an illustration of the COL) \cite{love2023tcnn}. These layers act as regularization mechanisms, improving generalization and reducing overfitting. TCNNs extend to video data using the tangent bundle of the Klein bottle to parameterize video filters.

For CF and KF layers, the convolutional filters are fixed to a discretization of the primary circle or the Klein bottle, respectively, embedded into the space of image patches as described in \cite{love2023tcnn}. These weights are not updated during training. For COL and KOL layers, the input and output slices are indexed by a discretization of either the primary circle or the Klein bottle. A metric induced by the embedding is used to enforce locality: connections between slices that are far apart on the manifold are set to zero, while connections between nearby slices are initialized randomly. The nonzero weights in COL and KOL layers are trained with standard neural network optimization methods. Compared to traditional convolutional layers, described in Section \ref{sec:CNN},  CF, KF, COL, and KOL layers compute the convolutional feature map in the same manner as ordinary convolutions, that is, having a sliding window kernel computing dot products, except these topological layers consider embeddings of particular manifolds.  

Formally, the embedding of the Klein bottle $\mathcal{K}$, where the Klein bottle is the 2-dimensional manifold given as a quotient by $(\theta_{1}, \theta_{2}) \sim (\theta_{1} + 2k\pi, \theta_{2} + 2l\pi)$ for $k, l \in \mathbb{Z}$ and $(\theta_{1}, \theta_{2}) \sim (\theta_{1} + \pi, - \theta_{2})$, into the space of functions on $[-1, 1]^{2}$ is given by 
\begin{equation}
    F_{\mathcal{K}}(\theta_{1}, \theta_{2})(x,y) = sin(\theta_{2})(cos(\theta_{1})x + sin(\theta_{1})y) + cos(\theta_{2}) Q (cos(\theta_{1})x + sin(\theta_{1})y),
\end{equation}
where $Q(\alpha) = 2\alpha^{2} - 1$. The primary circle is given as the subset of the Klein bottle $S^{1} = \{ (\theta, \frac{\pi}{2}) \in \mathcal{K}\}$. The embedding of the primary circle into the space of functions on $[-1, 1]^{2}$ is given by
\begin{equation}
    F_{S^{1}}(\theta)(x,y) = F_{\mathcal{K}}(\theta, \frac{\pi}{2})(x,y) = cos(\theta)x + sin(\theta)y.
\end{equation}


\textbf{Applications:} TCNNs have been shown to outperform CNNs in terms of accuracy, learning speed, and generalization across diverse datasets and tasks, and can be used in image and video classification settings, with potential extensions to 3D imaging and fMRI data. They are particularly effective in scenarios with limited data or high noise and generalize better across datasets with stylistic differences. Moreover, TCNNs are more interpretable, as their activations correspond to meaningful features like edges and lines.
\subsection{Bayesian Topological Convolutional Neural Networks}\label{sec:btcnn}

Building on the probabilistic weight formulation of BNNs, the Bayesian Topological Convolutional Neural Network (BTCNN), introduced in \cite{dayton2025bayesiantopologicalconvolutionalneural}, retains the probabilistic treatments of the model parameters while imposing topology-aware learning. The BTCNN addresses overfitting that frequently plagues CNNs, enhances model calibration, and enables uncertainty quantification. The BTCNN uses topology results from \cite{love2020tcnn, love2023tcnn} to parameterize convolutional filters on the primary circle and Klein bottle, yielding improved model performance over CNN, BNN, and TCNN. On an image classification task with both full-sized and reduced training sets, the BTCNN has been shown to outperform CNN, BNN, and TCNN in terms of accuracy and produce comparative calibration results to the BNN. Thus, the BTCNN is a well-equipped model for handling data-starvation settings and noisy, perturbed data. The BTCNN also showed to produce more appropriate uncertainty for test samples both in- and out-of-distribution. The hybrid nature of the BTCNN enables topological feature extraction and uncertainty quantification while increasing model performance, showing it to be a powerful model for natural image processing. The balance between accuracy and appropriate uncertainty is critical for high-stakes decision making.

\begin{figure}[ht]
    \centering
    \includegraphics[width=0.95\linewidth]{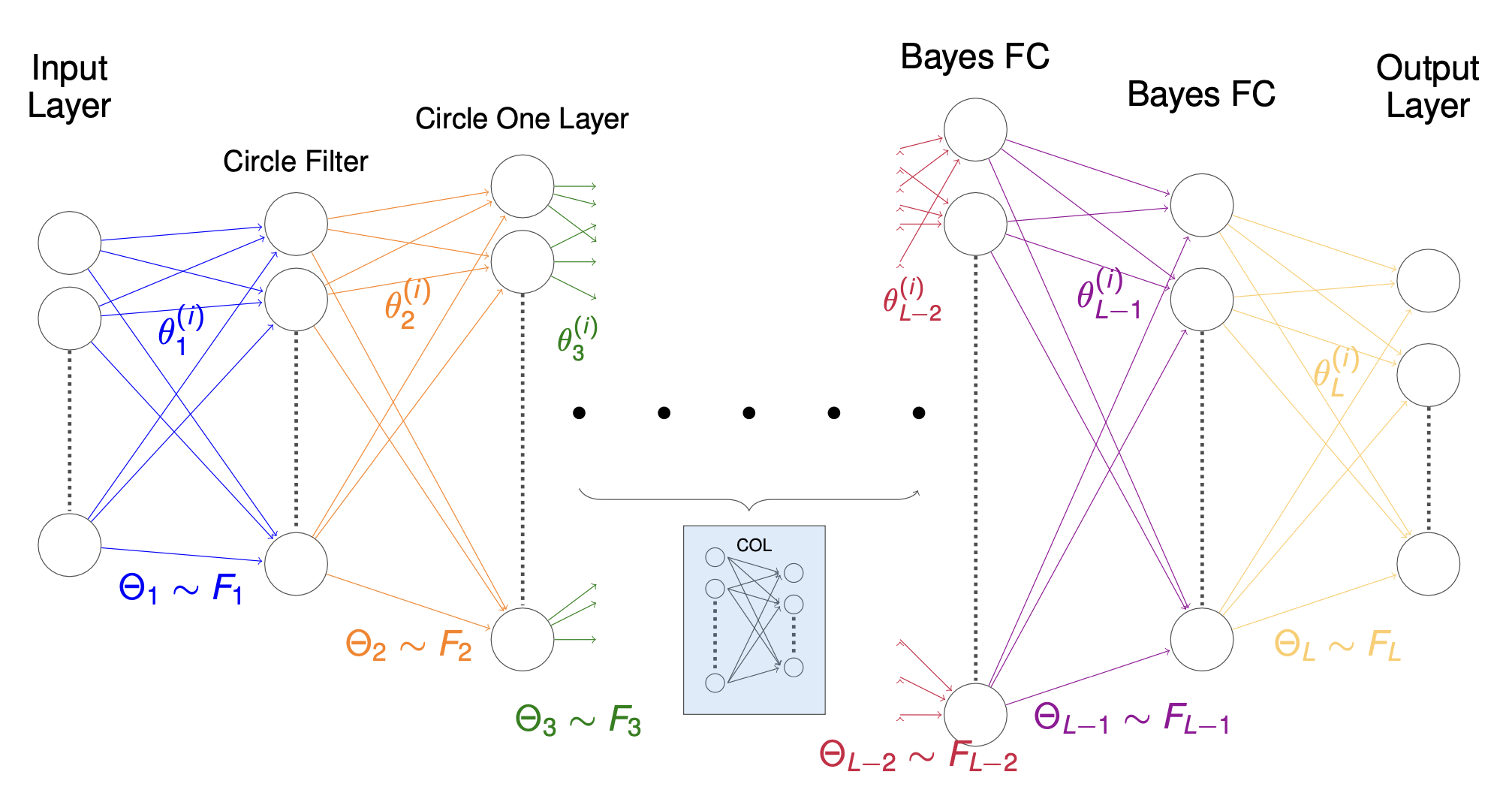} 
    \caption{A Bayesian topological convolutional neural network architecture where the first layer is a circle filter followed by max pooling, then an arbitrary number of circle one layers, each followed by max pooling, lastly, there are two Bayesian fully connected layers. Each layer $l$ has weights that follow some learned distribution, that is $\Theta_{l} \sim F_{l}$.}
    \label{fig:btcnn_arch}
\end{figure}

\textbf{Applications:} The discussed neural network techniques for images can be extended to process video inputs. Videos can be understood as sequences of images, where each frame captures spatial information and the progression of frames over time encodes motion and temporal dynamics. While the neural network techniques discussed in Sections \ref{sec:CNN}, \ref{sec:transformers}, \ref{sec:bnn}, \ref{sec:tcnn}, and \ref{sec:btcnn} were originally designed for still image analysis, their ability to learn spatial hierarchies makes them a strong foundation for processing video data when combined with techniques that model temporal information. Extending CNNs to video enables a wide range of applications, including video classification, activity recognition, and video summarization. To accomplish this, researchers have developed a wide range of techniques that extend CNNs to video \cite{diba2017temporal3dconvnetsnew, jiang2017modelingmultimodalclueshybrid, tran2018closerlookspatiotemporalconvolutions, ng2015shortsnippetsdeepnetworks}. These extensions make CNNs effective for analyzing complex, dynamic scenes. In military or defense settings, such techniques could be applied to automated surveillance, detection of suspicious activities, and recognition of tactical events from drone or satellite footage, offering enhanced situational awareness and decision-making support.

\subsection{Simplicial Neural Networks}

Simplicial neural networks (SNNs) introduced by Ebli et al. \cite{ebli2020simplicial} generalize neural networks to simplicial complexes and enable higher-order learning.  The core formulations include simplicial neural network \cite{ebli2020simplicial}, message passing simplicial network \cite{bodnar2021simplicial}, simplicial convolutional neural network \cite{yang2022simplicial}, and simplicial convolutional recurrent neural network \cite{MitchellSpikeDecoding}. Collectively, these works formalize Hodge-theoretic convolutions, inter-simplicial couplings, and provable expressivity, establishing SNNs as a practical tool for scientific and networked data.

Variants of simplicial neural networks deliver concrete gains across disciplines such as citation networks \cite{ebli2020simplicial,bodnar2021simplicial,yang2022simplicial} and trajectory prediction \cite{yang2023convolutional,pmlr-v139-roddenberry21a}. For example, \cite{pmlr-v139-roddenberry21a} used a simplicial neural network for predicting trajectories on synthetic data and real ocean-drifter tracks and grid-map navigation with holes or obstacles (e.g., Madagascar and Berlin map cases). The architecture showed a stronger generalization to unseen trajectories than baselines because it is permutation- and orientation-equivariant and “simplicial-aware”. 

To make use of algebraic methods, we assign an orientation to each simplex by ordering its vertices. For a $k-$simplex $\sigma_k=[v_0,\dots,v_k]$, the boundary operator is $\partial_k\sigma_k=\sum_{i=0}^{k}(-1)^i[v_0,\dots,\hat v_i,\dots,v_k]$,  where the alternating signs ensure consistency in the way simplices combine or cancel, and with it a coherent system of signs that records how parts agree or cancel when assembled. These sign conventions can be expressed in the incidence matrices $\mathbf{B}_k\in\mathbb{R}^{N_{k-1}\times N_k}$, which record the relationships between $(k-1)$- and $k$-simplices, where $N_{k-1}$ and $N_{k}$ are the number of $(k-1)$ and $k$ simplices. In this framework, the coboundary operator is represented by  $\delta^{k-1}=\mathbf{B}_k$, while the transpose $\mathbf{B}_k^\top$ corresponds to the adjoint operation, mapping in the opposite direction.  Local interactions between simplices are coupled globally by the $k$-dimensional Hodge Laplacian $\mathbf{L}_k = \mathbf{B}_k^{\top}\mathbf{B}_k + \mathbf{B}_{k+1}\mathbf{B}_{k+1}^{\top}$, whose action smooths signals across simplices, while its kernel $\ker \mathbf{L}_k $ identifies harmonic representatives of cohomology classes.

\subsubsection{Message Passing Simplicial Networks}\label{sec:message_simplicial}
The message passing simplicial network \cite{bodnar2021simplicial} updates $k$-simplex features by exchanging messages from faces through $\mathbf{A}_k^{\downarrow}=\mathbf{B}_k^{\top}\mathbf{B}_k-\operatorname{diag}(\mathbf{B}_k^{\top}\mathbf{B}_k)$,  from cofaces through $\mathbf{A}_k^{\uparrow}=\mathbf{B}_{k+1}\mathbf{B}_{k+1}^{\top}-\operatorname{diag}(\mathbf{B}_{k+1}\mathbf{B}_{k+1}^{\top})$,  and across dimensions via the boundary and coboundary maps. 
Given features $\mathbf{H}_{k}\in\mathbb{R}^{n_k\times d}$ on $k$-simplices, one propagation layer is
\begin{equation}
    \mathbf{H}^{(l+1)}_{k} =
\sigma\!\left(
\mathbf{W}_s \mathbf{X}^{(l)}_{k}
+ \mathbf{W}_{\downarrow} \mathbf{A}_k^{\downarrow} \mathbf{H}^{(l)}_k
+ \mathbf{W}_{\uparrow} \mathbf{A}_k^{\uparrow} \mathbf{H}^{(l)}_k
+ \mathbf{W}_{\delta} \mathbf{B}_k \mathbf{H}^{(l)}_{k-1}
+ \mathbf{W}_{\partial} \mathbf{B}_{k+1}^{\top} \mathbf{H}^{(l)}_{k+1}
\right),
\end{equation}
where $\sigma$ is a pointwise nonlinear activation function, such as ReLU, and $\mathbf{W}_s, \mathbf{W}_{\downarrow}, \mathbf{W}_{\uparrow}, \mathbf{W}_{\delta}, \text{ and } \mathbf{W}_{\partial}$ are trainable weight matrices.

\textbf{Applications:} Message passing simplicial network achieves state-of-the-art accuracy on higher-order link prediction and complex classification benchmarks on both synthetic and real-world simplicial complex datasets \cite{bodnar2021simplicial}.

\subsubsection{Simplicial Convolutional Neural Networks}
Although message passing simplicial networks work well for explicit cross-dimensional message flow, they do not impose a global spectral filtering prior. In settings with noisy or diffuse signals, simplicial convolutional neural networks, introduced by \cite{yang2022simplicial} offer polynomial filtering for denoising, imputation, and scalable regression on large complexes. They extended CNN to capture higher-order relations in which learning occurs on $k$-simplicial signals defined over simplicial complexes.

Let $\mathbf{X}^{(k)} \in \mathbb{R}^{N_k \times d_{\text{in}}}$ denote the input feature signal supported on $k$-simplices and let $\mathbf{Y}^{(k)} \in \mathbb{R}^{N_k \times d_{\text{out}}}$ denote the output (filtered) signal on the same $k$-simplices. The simplicial convolution is implemented as a polynomial filter in the lower and upper Hodge Laplacians,
\begin{equation}
    \mathbf{Y}^{(k)} \;=\; \sum_{i=0}^{p} a_i \big(L_{\downarrow}^{(k)}\big)^{i} \mathbf{X}^{(k)}
\;+\;
\sum_{j=0}^{q} b_j \big(L_{\uparrow}^{(k)}\big)^{j} \mathbf{X}^{(k)},
\end{equation}
where $a_i$ and $b_j$ are learnable scalar coefficients (trainable parameters of the filter), $\mathbf{L}_{\downarrow}^{(k)} = \mathbf{B}_k^{\top}\mathbf{B}_k$ is the lower Laplacian that propagates through shared $(k\!-\!1)$-faces, and $\mathbf{L}_{\uparrow}^{(k)} = \mathbf{B}_{k+1}\mathbf{B}_{k+1}^{\top}$ is the upper Laplacian that propagates through shared $(k\!+\!1)$-cofaces. The incidence matrices $\mathbf{B}_k$ encode signed boundary relations and ensure permutation- and orientation-equivariance.

\textbf{Applications:} The model was evaluated on citation imputation over a coauthorship simplicial complex to predicting missing interactions among coauthor groups. It reported gains over baseline models.

\subsubsection{Simplicial Convolutional RNN}

To extend simplicial convolutional neural networks to work on temporal data,  \cite{MitchellSpikeDecoding} introduced a simplicial convolutional recurrent neural network (SCRNN) that applies topological deep learning with unsupervised discovery of simplicial complexes for neural spike decoding. The brain encodes spatial information with head direction and place cells for orientation and environmental navigation. These cells fire in ensembles, and the signals gathered from these firings can be captured and decoded to recover the head direction and location. In this framework, neural spike trains are preprocessed so that each time bin yields a simplicial complex representing co-activation patterns, simplicial convolutions extract higher-order structure per time step, and a recurrent module models temporal dependencies to predict head direction and location.

The neural spike data is first represented as a raster plot, then, the data is binned and converted to a spike count matrix, $\mathbf{A} \in \mathbb{R}^{N \times N_{b}}$, where there are $N$ neurons and $N_{b}$ non-overlapping bins each with length $t_{b}$. Element $\mathbf{A}_{ij}$ is equal to the number of spikes from neuron $i$ in bin $j$. 

Next, the data is converted to a binarized matrix using a row-wise thresholding procedure. Consider a row of matrix $\mathbf{A}$ with elements $\{ a_{\ell} \}_{\ell = 1}^{N_{b}}$ where the elements are ordered from highest to lowest. For some $ p \in (0,1]$, we select $\{ a_{\ell} \}_{\ell = 1}^{m^{\ast}}$, where $m_{\ast} = \argmin_{1 \leq m \leq N_{b}} \Biggl\{ \sum_{\ell = 1}^{m} a_{\ell} \Bigg|  \sum_{\ell = 1}^{m} a_{\ell} \geq p \cdot\sum_{\ell = 1}^{N_{b}} a_{\ell} \Biggr\} $. Then, the entries selected in $\{ a_{\ell} \}_{\ell = 1}^{m^{\ast}}$ are set to 1 and those not selected are set to 0. This process is applied to all rows of $A$. In the resulting binarized matrix, where nodes are connected, representing brain neurons that have fired together.

 Lastly, the binarized matrix is converted to a simplicial complex. For each column of the binarized matrix, each neuron active in the column is connected with the appropriate dimension simplicial complex. A degree-$D$ simplicial filter, for a simplicial convolution (SC), is defined by 
\begin{equation}
     \mathbf{P}_{k} = \mathbf{W}_{0} \mathbf{I} + \sum_{i=1}^{D}\mathbf{W}_{i}(\mathbf{B}_{k}^{T} \mathbf{B}_{k})^{i} + \sum_{i=1}^{D}\mathbf{W}_{i+D}(\mathbf{B}_{k+1} \mathbf{B}_{k+1}^{T})^{i} \in \mathbb{R}^{N_{k} \times N_{k}},
\end{equation}
 where $k$ is the dimension of the simplices. The simplical filter is made up of weights $\mathbf{W} = \{ \mathbf{W}_{i}\}^{2D}_{i=0}$. Considerations are made to prevent exponential growth of the parameters with respect to the number of filters and the number of layers. The output of the SC layers, considered at a desired number of consecutive time bins, is taken as the input of a multilayer RNN. The outputs of the RNN are taken as the model's prediction.  

\textbf{Applications:} Simplicial complexes describe multiway relationships, whereas graph neural networks focus on pairwise connectivity between neurons. Brain neurons utilize connectivity between multiple neurons; thus, the authors claim that simplicial complexes are better suited for the task. The SCRNN has been shown to outperform a feedforward neural network, a recurrent neural network, and a graph neural network in predicting head direction and grid activation.

\subsection{Cell Complex Neural Networks}

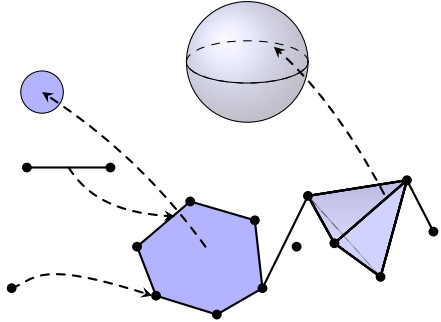
\begin{figure}[ht]
    \centering
    \begin{tikzpicture}[line cap=round,line join=round]

\tikzset{
  v/.style={circle,fill=black,inner sep=1.3pt},
  cell/.style={fill=blue!30,draw=black,thick},
  skel/.style={draw=black, thick},
  map/.style={dashed,->,>=stealth,thick}
}

\fill (0,1.6) node[v]{} (1.1,1.6) node[v]{};
\draw[skel] (0,1.6)--(1.1,1.6);

\fill (-0.2,0) node[v]{};
\draw[map] (-0.2,0) .. controls (0.1,0.1) and (0.1,0.4) .. (1.65,-0.1);

\path (1.7,-0.1) coordinate (a1)
      (2.5,-0.35) coordinate (a2)
      (3.1,0.0) coordinate (a3)
      (3.0,0.9) coordinate (a4)
      (2.15,1.15) coordinate (a5)
      (1.45,0.55) coordinate (a6);
\filldraw[cell] (a1)--(a2)--(a3)--(a4)--(a5)--(a6)--cycle;
\foreach \p in {a1,a2,a3,a4,a5,a6}{\fill (\p) node[v]{};}

\fill[blue!30] (0.2,2.6) circle (0.28);
\draw (0.2,2.6) circle (0.28);
\coordinate (pin) at (2.35,0.55); 
\draw[map] (pin) .. controls (1.6,1.6) and (0.8,2.2) .. (0.2,2.6);

\draw[map] (0.55,1.6) .. controls (0.8,1.2) and (1.2,1.0) .. (1.95,0.95);

\begin{scope}[shift={(4.35,0.85)}]
  \begin{scope}[tdplot_main_coords,scale=0.55]
    \def\R{1.6}
    \pgfmathsetmacro{\p}{\R/sqrt(3)}
    \coordinate (A) at (\p,\p,\p);
    \coordinate (B) at (-\p,-\p,\p);
    \coordinate (C) at (-\p,\p,-\p);
    \coordinate (D) at (\p,-\p,-\p);
    \shade[fill=blue!30,opacity=0.35] (A)--(B)--(C)--cycle;
    \filldraw[fill=blue!30,opacity=0.35] (A)--(B)--(D)--cycle;
    \filldraw[dashed][fill=blue!30,,opacity=0.35] (B)--(C)--(D)--cycle;
    \filldraw[fill=blue!30,opacity=0.35] (A)--(C)--(D)--cycle;
    \foreach \x in {A,B,C,D}{\path (\x) node[circle,fill=black,inner sep=1.3pt]{};
    \draw[skel] (C)--(B)  (C)--(D) (A)--(D) (C)--(A) (A)--(B);}
  \end{scope}
\end{scope}

\path (3.55,0.55) coordinate (b1)
      (5.35,0.75) coordinate (b2);
\draw[skel] (a3)--(B)  (A)--(b2) ;
\foreach \p in {b1,b2}{\fill (\p) node[v]{};}

\def\cx{2.9}\def\cy{3.0}\def\R{0.8}
\shade[ball color=blue!30,opacity=0.35] (\cx,\cy) circle (\R);
\draw (\cx,\cy) circle (\R);
\draw[map] (4.7,1.25) .. controls (4.2,2.2) and (3.7,2.8) .. (3.25,3.2);
\draw[dashed] (\cx,\cy) ellipse [x radius=\R, y radius=0.28];

  \clip (\cx-1.2*\R,\cy-1.2*\R) rectangle (\cx+1.2*\R,\cy); 
  \draw (\cx,\cy) ellipse [x radius=\R, y radius=0.28];

\end{tikzpicture}
    \caption{The homeomorphisms from closed balls to four cells of various dimensions, taken within a cell complex.} \label{fig:cell}
\end{figure}

Cell complex neural networks generalize simplicial complexes further by incorporating cells of arbitrary shapes and dimensions, allowing for the modeling of topologies that are not strictly simplex-based. This flexibility can capture more complex relational structures in data, such as irregular surfaces or multi-modal interactions \cite{bodnar2021cell, hajij2020cell}. Each cell is homeomorphic to an open ball of some dimension, and the boundary of each cell is attached to lower-dimensional cells, while the intersections of cells are contained in these boundaries. In particular, such structures are often formalized as CW complexes, where ``CW'' stands for closure-finite weak topology, meaning that each cell’s closure intersects only finitely many other cells and the topology is the weak topology determined by the cells. Examples include 2-cells that are squares or polygons, 3-cells that are cubes or prisms, and edges and vertices serving as 1-cells and 0-cells, respectively. See Figure \ref{fig:cell}.

In CW Networks \cite{bodnar2021cell}, a similar approach to message passing simplicial networks, discussed in Section \ref{sec:message_simplicial}, replaces simplices with cells and the simplex boundary $\mathbf{B}_k$ with the cellular incidence $\mathbf{D}_k$. The within-dimension adjacencies are $\mathbf{A}_k^{\downarrow}=|\mathbf{D}_k|^{\top}|\mathbf{D}_k|-\operatorname{diag}\!\big(|\mathbf{D}_k|^{\top}|\mathbf{D}_k|\big)$ and $\mathbf{A}_k^{\uparrow}=|\mathbf{D}_{k+1}||\mathbf{D}_{k+1}|^{\top}-\operatorname{diag}\!\big(|\mathbf{D}_{k+1}||\mathbf{D}_{k+1}|^{\top}\big)$, and the cellular Hodge Laplacian is $\mathbf{L}_k = \mathbf{D}_k^{\top}\mathbf{D}_k + \mathbf{D}_{k+1}\mathbf{D}_{k+1}^{\top}$.

\textbf{Applications:} The CW Networks framework is broadly applicable across domains where interactions occur among groups rather than merely between pairs. It outperforms or complements traditional graph neural networks by using higher-order topology for graph classification and regression tasks on both real-world molecular prediction problems and theoretical graph discrimination benchmarks. The main applications include molecular property prediction on standard chemistry benchmarks such as zinc, proteins, enzymes, where molecules or protein structures are modeled as cellular complexes to capture interactions beyond pairwise bonds. In synthetic graph datasets designed to test expressive power,CW Networks can distinguish graph structures that are indistinguishable from conventional GNNs.

\subsection{Sheaf Neural Networks}
A well-known challenge in GNNs is oversmoothing, where node representations become indistinguishable as the number of layers increases \cite{bodnar2022neural}. Oversmoothing-aware architectures are useful for long-range communication network analysis, ensuring that deep message-passing over many hops does not erase critical unit-specific distinctions in joint operations. Traditional GNNs excel primarily in homophilic settings, where neighboring nodes share similar attributes, and struggle with heterophilic(non-homophilic) data, where connected nodes are not necessarily similar, or higher-order relationships involving groups of nodes.

To overcome these constraints, sheaf neural network has been introduced by \cite{hansen2020sheaf}, extending conventional graph convolutions to richer, higher-order topologies and geometries. Neural sheaf diffusion \cite{bodnar2022neural} and the Bayesian sheaf neural network \cite{gillespie2025bayesian} demonstrated the power of sheaves in heterophilic graphs. Sheaf neural networks represent a significant leap forward in geometric deep learning, offering a powerful, interpretable, and uncertainty-aware framework for learning over complex data topologies, while also showing great promise for mission-critical, real-world applications.


Sheaf neural networks utilize the mathematical framework of cellular sheaves. A cellular sheaf on an undirected graph $G=(V,E)$ assigns a vector space $\mathcal{F}(v)$ to every node $v\in V$ and a vector space $\mathcal{F}(e)$ to every edge $e\in E$. For each incidence relation $v \inc e$ between a node and an edge, there is a linear map $\mathcal{F}_{v \inc e}: \mathcal{F}(v)\to \mathcal{F}(e)$.

In sheaf-theoretic language, the space $\mathcal{F}(v)$ is the stalk over $v$, and the maps $\mathcal{F}_{v \inc e}$ are restriction maps. From these local assignments, we form cochain spaces. The space of 0-cochains is the direct sum of vertex stalks, $C^0(G;\mathcal{F}):=\bigoplus_{v\in V} \mathcal{F}(v)$, while the space of 1-cochains is the direct sum of edge stalks, $C^1(G;\mathcal{F}):=\bigoplus_{e\in E} \mathcal{F}(e)$. Thus 0-cochains collect node-based data, and 1-cochains collect edge-based data. The coboundary map $\delta:C^0(G;\mathcal{F})\to C^1(G;\mathcal{F})$ provides the fundamental connection between these spaces. Given $\mathbf{x}\in C^0(G;\mathcal{F})$, the component at edge $e$ (with an arbitrarily chosen orientation from $u$ to $v$) is $(\delta \mathbf{x})_e = \mathcal{F}_{v \inc e}(\mathbf{x}_v) - \mathcal{F}_{u \inc e}(\mathbf{x}_u)$. The sheaf Laplacian is then defined as
\begin{equation}
    L_{\mathcal{F}} := \delta^\top \delta : C^0(G;\mathcal{F})\to C^0(G;\mathcal{F}),
\end{equation}
generalizing the standard graph Laplacian by embedding the richer algebraic structure of the sheaf into the operator. This construction enables topology-aware message passing that proves effective in heterophilic settings by preserving distinct features between neighbors, making the model suitable for scenarios involving complex inter-node dynamics.

\subsubsection{Neural Sheaf Diffusion}
Neural Sheaf Diffusion~\cite{bodnar2022neural} replaces updates of the layers of the sheaf neural network with a learned diffusion process on a sheaf Laplacian. Implements a continuous-time sheaf diffusion view that recovers GNNs as discretizations and presents practical end-to-end learning of sheaf parameters that yields strong empirical performance on citation networks. It adapts the message-passing framework of GCNs to the sheaf context by replacing adjacency-based propagation with sheaf-Laplacian diffusion that respects the restriction maps. A core update rule is
\begin{equation}
    \mathbf{H}^{(k+1)}
= \mathbf{H}^{(k)} - \sigma\!\Big(\,\Delta_{\mathcal{F}(k)} \big(\mathbf{I}_n \otimes \mathbf{W}^{(k,1)}\big) \mathbf{H}^{(k)} \mathbf{W}^{(t,2)}\Big),
\end{equation}
where $\mathbf{H}^{(k)}$ denotes the node feature matrix at iteration $k$, $\mathbf{I}_n\otimes \mathbf{W}^{(k,1)}$ applies a learnable transformation across all node stalks, and $\mathbf{W}^{(k,2)}$ is an additional learnable weight matrix acting after diffusion. The function $\sigma$ denotes a nonlinearity, and the normalized sheaf Laplacian is $\Delta_{\mathcal{F}(t)} = \mathbf{D}_F^{-\tfrac{1}{2}} \, L_{\mathcal{F}} \, \mathbf{D}_F^{-\tfrac{1}{2}}$, where $\mathbf{D}_F$ is the block-diagonal degree matrix with each block corresponding to a node stalk.

\textbf{Applications:} The framework demonstrates applications on node classification tasks across both homophilous and heterophilous datasets, including citation networks and synthetic graphs designed to test heterophily robustness. Empirical results show that neural sheaf diffusion consistently outperforms standard message-passing GNNs and state-of-the-art heterophilous models, highlighting the power of topological structures in improving graph representation learning.
\subsubsection{Bayesian Sheaf Neural Networks}
Despite their potential, sheaf neural networks remain susceptible to deep learning challenges such as overfitting, hyperparameter sensitivity, and limited noise resistance. To address these issues, Bayesian sheaf neural networks (BSNNs) \cite{gillespie2025bayesian} modeled the sheaf Laplacian as a latent random variable, whose distribution is learned from the input data using variational inference. 

BSNN evaluates semi-supervised node classification on real graphs where only a small fraction of nodes are labeled and many connected nodes differ in class. Each node has features and a few labels are revealed; the model learns to propagate information across edges using a sheaf-based message-passing scheme while treating key parameters probabilistically for calibrated uncertainty. The model learns the cellular sheaf $\mathcal{F}=\{\mathcal{F}_{u\inc e}\}$ by maximizing the ELBO as in section \ref{sec:bnn}
\begin{equation}
    \mathcal{L}(\phi,\theta)
= -\,\mathbb{E}_{q_\phi(F\,|\,X,y)}\big[\log p_\theta(y\mid X,\mathcal{F})\big]
+ \mathrm{KL}\!\big(q_\phi(\mathcal{F}\,|\,X,y)\,\|\,p(\mathcal{F})\big),
\end{equation}
with gradients computed via the Kingma–Welling \cite{kingma2013auto} reparameterization. Priors and variational families match the sheaf map type. For linear $\mathcal{F}_{u\inc e}\in\mathbb{R}^{d\times d}$ and diagonal $\mathcal{F}_{u\inc e}=\mathrm{diag}(d_{u\inc e})$, BSNN uses factorized Gaussians with affine reparameterization. For special-orthogonal $\mathcal{F}_{u\inc e}\in SO(d)$, it defines a Cayley distribution on $SO(d)$ obtained by mapping a reparameterized skew-symmetric matrix through the Cayley transform.
When a closed-form KL to the prior is unavailable, BSNN uses a Monte Carlo KL estimate. This yields differentiable samples for all types of maps and a single KL-regularized objective optimized by stochastic gradients.

\tikzstyle{arrow} = [thick,->,>=stealth]

\tikzstyle{sheaf-layer} = [rectangle, 
minimum width=1cm, 
minimum height=3cm, 
text centered, 
draw=skyblue, 
fill=skyblue!40]

\tikzstyle{input} = [rectangle, 
minimum width=1cm, 
minimum height=3cm, 
text centered, 
draw=gray, 
fill=lightgray!30]

\tikzstyle{output} = [rectangle, 
minimum width=3cm, 
minimum height=1cm, 
text centered, 
draw=blue!80, 
fill=blue!30]

\tikzstyle{distribution} = [rectangle, 
minimum width=3cm, 
minimum height=1.25cm, 
text centered, 
draw=blue!80, 
fill=blue!15]

\tikzstyle{mlp} = [rectangle, 
minimum width=1cm, 
minimum height=3cm, 
text centered, 
draw=orange!80, 
fill=orange!15]

\tikzstyle{sheaf-mlp} = [rectangle, 
minimum width=2.25cm, 
minimum height=1.5cm, 
text width=2cm,
text centered, 
draw=orange!80, 
fill=orange!15]

\tikzstyle{my-samples} = [rectangle, 
minimum width=1cm, 
minimum height=1cm,
text centered, 
draw=blue, 
fill=blue!15]

\begin{figure}[H]
\centering
\resizebox{0.7\textwidth}{!}{
    \begin{tikzpicture}[node distance=2cm]
    
    \node (input) [input] {\rotatebox{90}{Input Graph}};
    \node (mlp1) [mlp, xshift=2cm] {\rotatebox{90}{MLP}};
    
    \node (sheaf1) [sheaf-layer, xshift=4cm] {\rotatebox{90}{SheafConv}};
    \node (sheaf2) [sheaf-layer, xshift=5.5cm] {\rotatebox{90}{SheafConv}};
    \node (sheafdots) [xshift=7cm] {$\cdots$};
    \node (sheaf3) [sheaf-layer, xshift=8.5cm] {\rotatebox{90}{SheafConv}};
   
    \node (mlp2) [mlp, xshift=10.5cm] {\rotatebox{90}{Linear Layer}};
    \node (output) [input, xshift=12.5cm] {\rotatebox{90}{Output}};
    
    \node (F1) [my-samples, xshift=4cm, yshift=2.5cm] {$\mathcal{F}_1$};
    \node (F2) [my-samples, xshift=5.5cm, yshift=2.5cm] {$\mathcal{F}_2$};
    \node (Fdots) [xshift=7cm, yshift=2.5cm] {$\cdots$};
    \node (F3) [my-samples, xshift=8.5cm, yshift=2.5cm] {$\mathcal{F}_L$};
    
    \node (dist) [distribution, xshift=6.25cm, yshift=4.75cm] {Sheaf Distribution};
    
    \node (sheaf-mlp) [sheaf-mlp, xshift=2cm, yshift=4.75cm] {Linear Layer + Concat. + MLP};
    
    \draw [arrow] (input) -- (mlp1);
    \draw [arrow] (mlp1) -- (sheaf1);
\draw [arrow] (sheaf1.east) -- (sheaf2.west);
\draw [arrow] (sheaf2.east) -- (sheafdots);
\draw [arrow] (sheafdots) -- (sheaf3.west);
    
    \draw [arrow] (sheaf3) -- (mlp2);
    \draw [arrow] (mlp2) -- (output);
    
    \draw [arrow] (F1) -- (sheaf1);
    \draw [arrow] (F2) -- (sheaf2);
    \draw [arrow] (F3) -- (sheaf3);
    
    \draw[dashed, arrow] (dist.south) |- ($(dist.south) - (0,0.5cm)$) -| (F1.north);
    \draw[dashed, arrow] (dist.south) |- ($(dist.south) - (0,0.5cm)$) -| (F2.north);
    \draw[dashed, arrow] (dist.south) |- ($(dist.south) - (0,0.5cm)$) -| (F3.north);
    
    \draw [arrow] (input) |- (sheaf-mlp);
    \draw [arrow] (sheaf-mlp) -- (dist);
    
    \end{tikzpicture}
}
\caption{Visual representation of the BSNN architecture. The upper portion of the diagram illustrates the variational sheaf learning mechanism, with dashed arrows indicating sampling from the sheaf distribution.}
\label{fig:architecture}
\end{figure}
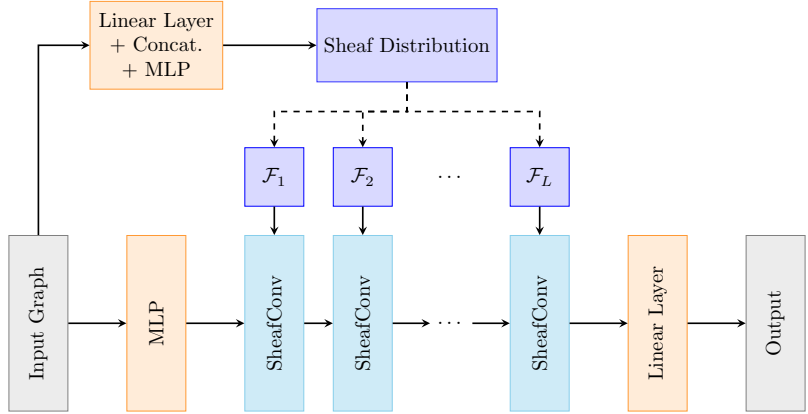

\textbf{Applications:} The Bayesian Sheaf Neural Network framework is applied to node classification tasks across both homophilous and heterophilous graph datasets, including citation networks and synthetic graphs designed to test robustness to label diversity and structural heterogeneity. Performance is assessed using accuracy and calibration metrics, demonstrating improved reliability in low-label regimes and on label-diverse networks that resemble real-world web, social, and citation settings. The probabilistic formulation of BSNN enables the model to quantify epistemic uncertainty and maintain expressive power in the presence of heterophily, making it well-suited for applications involving sensor networks, agent-based systems, and other domains with non-homogeneous relational structures.
\subsection{Topological Functional Units}


Topological Functional Units (ToFU) uses a PD dissimilarity function as its nonlinear activation function, thus utilizing the topology of the data for machine learning tasks \cite{OballeTOFU}. This technique learns parameters in PD space that help it topologically distinguish inputs. Traditional artificial neural networks (ANNs) rely on data augmentation, such as translations and rotations, to handle variability in the input. In contrast, ToFU learns features that are invariant to rigid transformations of the input by directly leveraging the persistent homology of the data. 

Let $\mathcal{D} = \{ \mathbf{p}_{n} \}^{N}_{n=1}$, where $\mathbf{p}_{i} \in \mathbb{R}^{2}$ $\forall i=1, \dots, N$,  denote a persistence diagram with an arbitrary ordering on its points. The ToFU layer is defined as $\phi_{\mathbf{\mathcal{D}}}: \mathbb{R}^{d} \rightarrow \mathbb{R}$ parameterized by $\mathbf{\mathcal{D}}$ with transformation $T(\mathbf{x}) := \mathcal{D}_{\mathbf{x}}$ and whose activation is given by $\sigma_{\mathcal{\mathbf{D}}} (\mathbf{D}_{\mathbf{x}}) = \frac{1}{2} m (\mathcal{D}_{\mathbf{x}}; \mathbf{\mathcal{D}})$, where $m$ is the utilized PD dissimilarity function, the minimal-cost matching function, defined by 
\begin{equation}
    m(\mathbf{\mathcal{D'}}; \mathbf{\mathcal{D}}) = \min_{\gamma \in \Pi} \sum_{\mathbf{p}\in \mathcal{D}} \|\mathbf{p} - \gamma(\mathbf{p}) \|_{2}^{2},
\end{equation}
where $\mathcal{D}$ and $\mathcal{D'}$ are PDs, $|\mathcal{D'}| \geq | \mathcal{D}|$, $| \cdot|$ being cardinality, and $\Pi$ being the set of injections from $\mathcal{D}$ to $\mathcal{D'}$.

\textbf{Applications:} In a classification task of discrete-time autoregressive signals, ToFU has been shown to outperform convolutional neural networks and produce comparable results to a model using spectral features. When used in a variational autoencoder, ToFU enables a topologically interpretable latent space without degrading reconstruction quality. ToFU offers a powerful technique for topology-grounded interpretability and task-oriented performance, particularly in settings where robustness to transformations and topological insight are essential. 
    
    \section*{Acknowledgments}
This work was supported in part by the National Science Foundation under grants DMS-2012609, DMR-2309083, DRL-2314155, DMS-1821241, MCB-1715794, and DGE-2152168; the U.S. Army DEVCOM Army Research Laboratory (ARL) under contracts W911NF-22-2-0143, W911NF-21-2-0186, and W911NF-19-2-0328; the U.S. Army Research Office (ARO) under contracts W911NF-21-1-0094 and W911NF-17-1-0313; and Thor Industries/ARL under contract W911NF-17-2-0141.

    \nocite{*} 
    \printbibliography
\end{document}